\documentclass{article} 
\usepackage[preprint]{colm2026_conference}

\usepackage{microtype}
\usepackage{hyperref}
\usepackage{url}
\usepackage{booktabs}

\usepackage{lineno}

\usepackage{amsmath}
\usepackage{amssymb}
\usepackage{mathtools}
\usepackage{amsthm}

\theoremstyle{plain}

\theoremstyle{definition}

\theoremstyle{remark}

\RequirePackage{algorithm}
\RequirePackage{algorithmic}

\usepackage{wrapfig}
\usepackage{amssymb}
\usepackage{enumitem}
\usepackage{etoolbox}
\usepackage[dvipsnames]{xcolor}
\usepackage{graphicx}
\usepackage[capitalize,noabbrev,nameinlink]{cleveref}
\usepackage[breakable]{tcolorbox}
\usepackage{fancyhdr}
\usepackage{subcaption}
\usepackage{wrapfig}
\usepackage{ifmtarg}
\usepackage{lipsum}

\definecolor{lightgray}{gray}{0.90}

\definecolor{cornflowerblue}{rgb}{0.39, 0.58, 0.93}
\definecolor{cornflowerscomplement}{rgb}{0.93, 0.74, 0.39}
\definecolor{cornflowersanalogouspurple}{rgb}{0.47, 0.39, 0.93}
\hypersetup{
    colorlinks=true,
    linkcolor=cornflowerblue,
    filecolor=magenta,      
    urlcolor=cornflowersanalogouspurple,
    citecolor=cornflowerblue,
    pdftitle={Multi-Token Prediction via Self-Distillation},
    }

\title{Multi-Token Prediction via Self-Distillation}

\author{
\textbf{John Kirchenbauer$^1$\thanks{Correspondence to \texttt{<jkirchen@umd.edu>}. Code available at: \href{https://github.com/jwkirchenbauer/mtp-lm}{\texttt{github.com/jwkirchenbauer/mtp-lm}}. Checkpoints available at: \href{https://hf.co/collections/tomg-group-umd/mtp-lm}{\texttt{hf.co/collections/tomg-group-umd/mtp-lm}}.}
, Abhimanyu Hans$^1$} \\
\textbf{Brian Bartoldson$^2$,Micah Goldblum$^3$, Ashwinee Panda$^{1,4}$, Tom Goldstein$^1$} \\
\\
$^1$University of Maryland\\
$^2$Lawrence Livermore National Laboratory
$^3$Columbia University
$^4$TogetherAI
}

\begin{document}

\ifcolmsubmission
\linenumbers
\fi

\maketitle

\begin{abstract}
Existing techniques for accelerating language model inference, such as speculative decoding, require training auxiliary speculator models and building and deploying complex inference pipelines. We consider a new approach for converting a pretrained autoregressive language model from a slow single next token prediction model into a fast standalone multi-token prediction model using a simple online distillation objective. The final model retains the exact same implementation as the pretrained initial checkpoint and is deployable without the addition of any auxiliary verifier or other specialized inference code. Our method produces models that decode more than $3\times$ faster at $<5\%$ drop in accuracy on GSM8K relative to the single token decoding performance of the same checkpoint.
\end{abstract}

\section{Introduction}

Standard language models generate text by predicting tokens one-at-a-time, resulting in very slow throughput when generating many tokens.  This problem is exacerbated in reasoning models, as they often use thousands of tokens during a chain of thought even when they will ultimately produce a short final response, resulting in a sluggish user experience. To address this issue, we consider an approach for training a language model (LM) to produce multiple tokens in a single forward pass, enabling it to generate spans of text with significantly reduced latency and inference cost. 

\begin{figure}[h!]
    \centering
    \includegraphics[width=\columnwidth]{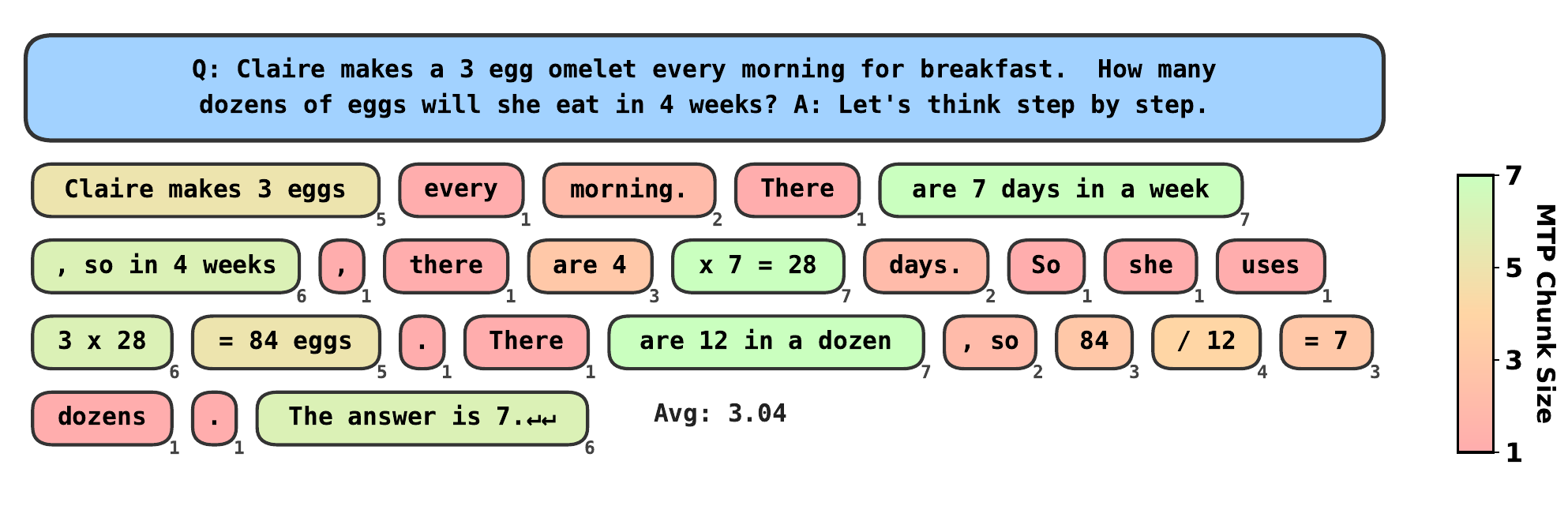}
    \caption{Example response to a GSM8K test question from our Qwen3-4B-Instruct-2507 based multi-token prediction model. Decoding is performed using a confidence-adaptive strategy with a threshold of 90\% with no secondary verification step. Each colored block corresponds to a chunk of tokens produced during a single forward pass and is annotated with its size in tokens which ranges from 1 to 7 in this example. The average chunk size over the entire generation is 3.04.}
    \label{fig:explainer-example-chunks}
\end{figure}
Existing work on multi-token prediction (MTP) treats it as merely instrumental to inference acceleration. Prior work primarily trains MTP heads as speculators whose predictions must be verified in a secondary stage, or simply uses MTP as an auxiliary objective to try and improve standard next-token prediction (NTP) decoding. \textbf{In this work, we pursue multi-token prediction as a first class, standalone inference paradigm.}

Our approach is motivated by the observation that the standard offline cross-entropy loss has fundamental limitations that make it difficult or even impossible to learn to generate long, grammatical spans of text. This problem is most clearly illustrated using an example. Consider a model trained on the two sentences ``The zookeeper fed the panda bamboo'' and ``The zookeeper fed the lion meat''. If we predict two tokens conditioned on the prefix ``The zookeeper fed the'', then an optimal model should assign equal probability to $\{panda, lion\}$ for the next token, followed by equal probabilities for $\{bamboo, meat\}$.  If we sample both words at the same time, then half the time our zookeeper  feeds the panda meat or the lion bamboo. In the abstract, a naive multi-token predicting LM would sample each word independently and ignore the correlations in the joint distribution between tokens.  For MTP to succeed, we need a training loss, and a sampler, that cares about the {\em joint} distribution.

In this paper, we propose an RL-inspired training paradigm in which a student model generates a span of simultaneous token predictions. To avoid the pitfalls of the standard offline objective, the student output is scored by an LM critic/teacher, rather than being scored against a known ground-truth token sequence. By comparing the student's predictions against the next-token suggestions made by the teacher, we produce an on-policy reward signal that enables the student to quickly improve the quality of its multi-token predictions.

After exploring how to implement this training objective in an efficient and scalable way, we turn our attention to inference time sampling strategies.  While our method is capable of producing blocks of $k$ tokens on every forward pass for $k$ as large as 16, we can avoid compromising accuracy while still achieving robust speedups using an adaptive strategy in which tokens are only kept if they have high confidence. This strategy naturally enables the model to produce multiple ``easy'' tokens on a forward pass, while focusing its costly single-token passes on ``hard'' tokens that require more computational effort.  This automatic moderation of the speed vs. accuracy tradeoff in various domains like grade school math, instruction following, and open-ended generation allows us to achieve between $2\times$ and $5\times$ acceleration with minimal impact on generation quality.

\section{Related Work}

\paragraph{Multi-token prediction.}\label{sec:rel-mtp}
Recent work explores architectural and training modifications for building LMs with multi-token-prediction abilities. \citet{samragh2025your} enhance coherence of predicted tokens by augmenting the unembedding layer with an MLP that integrates the previously sampled token's embedding, and further use MTP tokens as drafts in speculative decoding. \citet{cai2025fastmtp} finetune a single-layer MTP module via self-distillation with cross-entropy loss, observing that self-distilled data better aligns MTP predictions with baseline model outputs. Self-distilled training data has also been used in the Parallel Token Prediction models of \citet{draxler2025parallel} and the Jacobi Forcing Models of \citet{hu2025fast}. We primarily motivate MTP as a means to accelerate generation without significant quality loss, but other work treats MTP as a surrogate training objective component to improve next-token modeling \citep{liu2024deepseek}, or as a way to address failure modes in long-horizon planning by reducing reliance on teacher-forcing prefixes \citep{bachmann2024pitfalls,mahajan2025beyond}.

\paragraph{Speculative decoding.} \citet{leviathan2023fast}, \citet{chen2023accelerating}, and \citet{li2025eagle} show that a faster LM can sequentially produce draft tokens that a larger LM verifies in parallel, accepting only tokens the larger LM would itself generate---a lossless criterion that can still yield speedups. Some approaches use a dedicated MTP module as the speculator \citep{cai2024medusa}. However, real-world speedups have been questioned due to factors like multi-user batching and dataset sensitivity \citep{liu2025speculative}.

\paragraph{Language model distillation.} Teacher capabilities can be transferred to a student via knowledge distillation~\citep{kim2016sequence}. \citet{agarwal2024policy} perform on-policy distillation---using teacher feedback on student-generated sequences---to avoid distributional mismatch with static training data. \citet{zhou2023distillspec} apply distillation to align draft models with target models, improving speculative decoding acceptance rates.

\paragraph{Online RL and entropy minimization.}
\citet{wang2021tent} first showed generalization benefits of direct entropy minimization for image classifiers; modern studies demonstrate similar effects in LMs trained with reinforcement learning from verifiable rewards~\cite{agarwal2025unreasonable,shao2024deepseekmathpushinglimitsmathematical}. \citet{kang2025scalable} exploit the correlation between model confidence and correctness for best-of-$n$ sampling at test time, and \citet{fu2025deep} leverage it to re-weight rollouts during online and offline RL, observing consistent improvements. We build on these results by designing our MTP objective and inference strategies around the correspondence between low-entropy predictive distributions and generation quality in reasoning-intensive domains.

\section{Methodology}\label{sec:prelim}

In \cref{sec:prelim-ntp} through \cref{sec:prelim-mtpo2} we motivate and define our novel MTP LM training objective in an abstract manner and in \cref{sec:prelim-inference} we discuss how to perform inference using this kind of model. We also provide a formalization of how to perform a training step using our method in \cref{alg:singleshot-online-train-step}. Then in \cref{sec:impl} we discuss specific aspects of our method in more concrete, implementation focused terms with accompanying visualizations.

\subsection{Next Token Prediction (NTP)}\label{sec:prelim-ntp}
Let $\mathcal{V}$ be the set of $V$ tokens in the vocabulary of our LM, and consider an input sequence $X = (x_1,x_2,...x_N) \in \mathcal{V}^N$. A transformer LM $f_\theta(\cdot)$ parametrized by neural network parameters $\theta$ maps an input $X$ into a sequence of logit vectors $\{\ell_i\in \mathbb{R}^V$\}: 
\begin{equation*}
f_\theta(X) = (\ell_1,\ell_2,..\ell_N) \in \mathbb{R}^{N \times V}.
\end{equation*}
To produce tokens from these logits, we consider a \textit{readout} function $g: \mathbb{R}^{1 \times V} \to \mathcal{V}$ which could be:
\begin{align*}
g(\ell_i) = \operatorname{argmax}_j(\ell_{ij}) \quad\text{or}\quad g(\ell_i) = \operatorname{sample}(\operatorname{softmax}(\ell_i)).
\end{align*}
For brevity, in subsequent sections we refer to the $\operatorname{sample}(\operatorname{softmax}(\cdot))$ composition as just ``softmax'' when contrasting it with ``argmax''.

Let $Y$ be the ground truth next tokens at each position in $X$ such that $y_i = x_{i+1}$, and let the predicted next token distribution parametrized by the model be 
\begin{equation*}
P_\theta(x_{i+1}| x_{1:i}) := g(\ell_i) = \operatorname{sample}(\operatorname{softmax}(\ell_i)).
\end{equation*}
If the ground truth next tokens are represented as one-hot vectors denoted $y_i= \{0,1\}^{|\mathcal{V}|}$, then the cross-entropy training objective for a next token prediction language model is given as:
\begin{equation}\label{eq:ntp-ce-sum-log}
\mathcal{L}_{NTP} = - \frac{1}{N}\sum_{i=1}^{N} \log P_\theta(y_i | x_{1:i}).
\end{equation}

\subsection{Multi-Token Prediction (MTP)}\label{sec:prelim-mtp}

In our MTP model, the input $x$ contains prefix tokens, followed by a block of $k-1$ MTP ``mask'' tokens.
On a forward pass, the model predicts $k$ tokens, one at the final prefix position and $k-1$ tokens at each masked position. We can define readouts that operate over multiple logit vectors instead of one:
\begin{align*}
g(l_{i:i+k}) = \operatorname{argmax}_j(\ell_{i:i+k,j}) \quad\text{or}\quad g(l_{i:i+k}) = \operatorname{sample}(\operatorname{softmax}(\ell_{i:i+k})).
\end{align*}
Therefore, the multi-token prediction distribution parametrized by the model under the probabilistic readout would be
\begin{equation*}
P_\theta(x_{i:i+k} | x_{1:i}) := g(\ell_{i:i+k}) = \operatorname{softmax}(\ell_{i:i+k}).
\end{equation*}

A naive training loop for MTP can be constructed using a standard offline objective in which all $k$ token predictions are compared to $k$ ground-truth targets, computing the cross-entropy loss. However, in the next section we will introduce our proposal for a MTP training objective that avoids the potential pitfalls with such an approach.

\subsection{``Student Forced'' Online MTP}\label{sec:prelim-mtpo2}

We propose an RL-inspired alternate objective that rewards a coherent joint token distribution in a natural way. Consider a strong oracle NTP LM that will serve as our critic/teacher. Rather than use the ground truth next token distribution as the target distribution, we instead train the student MTP model, $P_{\theta_S}$, to predict $k$ next tokens that a strong \textit{teacher} NTP model, $P_{\theta_T}$, would assign a high likelihood to.  In our above example, an oracle LM would assign a low likelihood (high loss) to ``The zookeeper fed the lion bamboo,'' and so the student trained to minimize the teacher's criticism would not produce this sentence.

As the teacher is a standard NTP model, it does not parametrize the joint token distribution natively. However, by the chain rule of probability, we can use it to score the likelihood of a $k$ token sequence $y' \in \mathcal{V}^k$ proposed by the student. To materialize this sequence, we apply the deterministic readout to the student's logits so that $y':= (y_1,y_2,...y_k) = \operatorname{argmax}(\ell_{i:i+k})$. Then, we compute the likelihood of these generated tokens under the teacher conditioned on the ground truth prefix according to:
\begin{equation}\label{eq:teacher-decomp}
P_{\theta_T}(y' | x_{1:i}) = \prod_{j=1}^{k} P_{\theta_T}(y'_{j} | y'_{1:j-1} \oplus x_{1:i}).
\end{equation} 

Finally, we define our new \textit{online} objective as the KL divergence between the likelihood assigned to $y'$ by the student-forced teacher and the student's MTP distribution over its own generation $y'$. For a sequence $X$ broken up into $R$ multi-token-prediction regions (see \cref{fig:explainer-tok-masking}), each region $r$ with prefix $x_{1:i_r}$, the online training objective is:
\begin{align}\label{eq:mtpo2}
\mathcal{L}_{MTP} = - \dfrac{1}{R}\sum_{r=1}^{R} P_{\theta_T}(y' | x_{1:i_{r}}) \log P_{\theta_S}(y' | x_{1:i_{r}})
\end{align}
Note that even though it is non-zero in general, we omit the entropy term in the KL involving only $P_{\theta_T}$ as it is constant w.r.t. optimization of the student parameters $\theta_S$.

\paragraph{Advantages of the online approach.}
The objective presented in \cref{eq:mtpo2} has a few desirable properties. First, the student forcing operation means the objective is on-policy. We actually ``roll out'' a sequence of $k$ tokens from the student model and then compute a ``reward'' based on how likely a strong oracle model thinks they are. Note that, unlike in conventional RL for LMs, our roll outs comprise a single forward pass. Second, it is also stable across training steps and sequences in the full training corpus. Computing next token likelihoods in service of \cref{eq:teacher-decomp} requires that the teacher model be implemented using the softmax readout function. However, we are not \textit{sampling} from the teacher's distribution, only materializing it. This ensures that the feedback the teacher provides given the same ground truth prefix and student proposal $y' \oplus x_{1:i}$ is deterministic and non-contradictory. 

We expect that an online MTP objective such as ours allows the student model to more efficiently learn to produce a coherent MTP distribution than it would under an offline objective. Indeed, our ablations in \cref{fig:train-dynamics-flagship-l3-online-offline,fig:dynamics-flagship-l3-online-offline} show that this true in practice. Training in an offline manner where $P_{\theta_T}$ in \cref{eq:mtpo2} is replaced with delta distribution corresponding to the ground truth tokens in the training data yields inferior results even when the dataset comprises completions generated using the teacher model itself.

\paragraph{``Hard'' teacher distribution.}
Finally, to further enhance the clarity of our supervision signals, we also consider a version of \cref{eq:mtpo2} where the readout function applied to the teacher is also the argmax. This reduces $P_{\theta_T}(y'| x_{1:i_{r}})$ to a delta distribution with zero entropy. By construction, computing the loss using a ``hard'' version of the student-forced teacher's distribution requires that the student's entropy drop to zero to minimize it completely. While we do not achieve such full convergence in our experiments, we do observe the desired entropy reduction effect one would expect under this setting and adopt it for our main training runs.

\paragraph{Choice of teacher model.}

While \cref{eq:mtpo2} imposes no particular constraints on the exact model chosen as the teacher or the initialization of the student policy, we believe a natural choice exists. In the special case where $k=1$, during the first step of training, the online MTP objective reduces to a standard NTP knowledge distillation approach; the student need only predict a single token following a given prefix that is likely under the teacher. For this reason, we posit that initializing the student and the teacher using the same pretrained checkpoint, $\theta_T^0 = \theta_S^0$, ensures that the loss function is well behaved in the early stages of training. Under this initialization, at step 0 and $k=1$, the student and teacher's predictions are exactly the same and \cref{eq:mtpo2} is 0. Therefore, we will run all experiments under $\theta_T^0 = \theta_S^0$ initialization, but with $k \geq 2$ to allow the student to actually learn to perform accelerated inference.

\subsection{Inference}\label{sec:prelim-inference}


The training objective \cref{eq:mtpo2} uses the deterministic argmax readout function to materialize the student's $k$-token rollout $y'$. However, at inference time, we must also choose a method for sampling from the student. We note that in our simple MTP formulation, $P_{\theta_S}(y' | x_{1:i})$ does not explicitly parametrize the product distribution over $\mathcal{Y}^k$ and instead produces $P_{\theta_S}(y'_j | x_{1:i}), \quad j \in [1,..k]$. Therefore, under the softmax readout, the $k$ tokens would actually be sampled individually. If the distribution has any entropy---meaning the student is not perfectly certain about the individual tokens at each position---then the sampled tokens may not be mutually compatible.

In principle, this issue is resolved simply by using the greedy argmax readout function at test time. Because the model is trained using a deterministic argmax, a well fit MTP model should yield argmax tokens at each position that are compatible with their argmax neighbors. The only thing that matters for such a model is which token is assigned the largest logit value rather than the entropy of the model's entire output distribution. Therefore, we must determine empirically whether or not argmax tokens produced by a student model trained using~\cref{eq:mtpo2} are actually coherent in practice. Later, we will see that despite the fact that it is not strictly necessary, high generation coherence (low perplexity) appears to be closely correlated with high token confidence (low entropy) in our experiments, and we will exploit this fact to create a simple adaptive sampler.

\section{Implementation Details}\label{sec:impl}

\begin{figure*}[t!]
    \includegraphics[width=\textwidth, trim=1cm 4cm 0cm 2.5cm, clip]{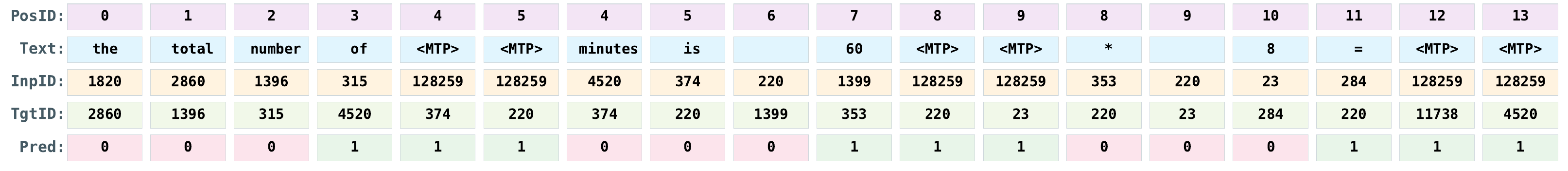}
    \caption{Visual depiction of how a piece of training text is tokenized and masked. In this example, the sequence length is 18, the $k$ value is 3, and the number of MTP regions $R$ is also 3. We show post-masking replication of the $k$ ground truth tokens corresponding to each MTP region and the corresponding position embedding adjustments required. Note that while the targets row (\texttt{TgtID}) is filled with ground truth tokens from the dataset for clarity, under our proposed online training objective, the targets (\texttt{Pred}) are based on the teacher model's feedback, not ground truth data. \textbf{The masking style and and input replication shown materializes many different MTP problems within a single sequence in parallel, increasing training efficiency.}}
    \label{fig:explainer-tok-masking}
\end{figure*}

In \cref{sec:prelim}, we introduced our method in an abstract manner. Now, in \cref{sec:impl-tok-mask} through \cref{sec:impl-confadapt} we provide details about the actual implementation of our training and inference scheme; complete details can be found in \cref{app:impl-exp-details} including a formal procedure in \cref{alg:singleshot-online-train-step}. \Cref{fig:explainer-tok-masking,fig:explainer-block-mask} are rendered in support of components of our implementation that are most succinctly described using visuals such as attention patterns and input tokenization and masking.

\subsection{Tokenization and Masking}\label{sec:impl-tok-mask}

The training objective proposed in \cref{eq:mtpo2} is computed over a specific position $i$ in a ground truth sequence $X$ and the next $k$ tokens that the student predicts to follow the prefix $x_{1:i}$. For any sequence $X$ of length $N$, there are $N-1$ unique prefixes that can be materialized as input for our online objective. To train efficiently, we need an attention masking scheme that allows us to supervise multiple prediction spans within a single sequence in parallel.

In \cref{fig:explainer-tok-masking}, we present a visual depiction of how input text is tokenized into a sequence, masked, and aligned for our model so that multiple supervised spans can be computed in parallel. To compute the $k$ token prediction $y'$ of the student at a given position in the tokenized input sequence, $k-1$ ``MTP masks'' are inserted directly following the last token of the prefix that will be provided as input to the model for MTP problem beginning at position $i$.

Another important detail shown in \cref{fig:explainer-tok-masking} is the appearance of ground truth tokens following the first MTP region. Suppose we make MTP predictions at position $i$, followed by another prediction at position $i'>i.$ This later prediction must have access to all ground truth tokens in the span $x_{0:i'}$, which includes the tokens $x_{i:i'}$ between $i$ and $i'$.  We place the ground truth tokens $x_{i:i'}$ immediately after the first MTP region. Then we also modify the position embeddings (e.g. RoPE features) so these tokens have their original position features (ignoring the presence of the MTP mask tokens) applied to them, though they now appear later in the context window.

\subsection{Blocked Attention}\label{sec:impl-block}

\begin{figure}[t!]
    \begin{subfigure}[b]{0.24\textwidth}
    \includegraphics[width=\textwidth]{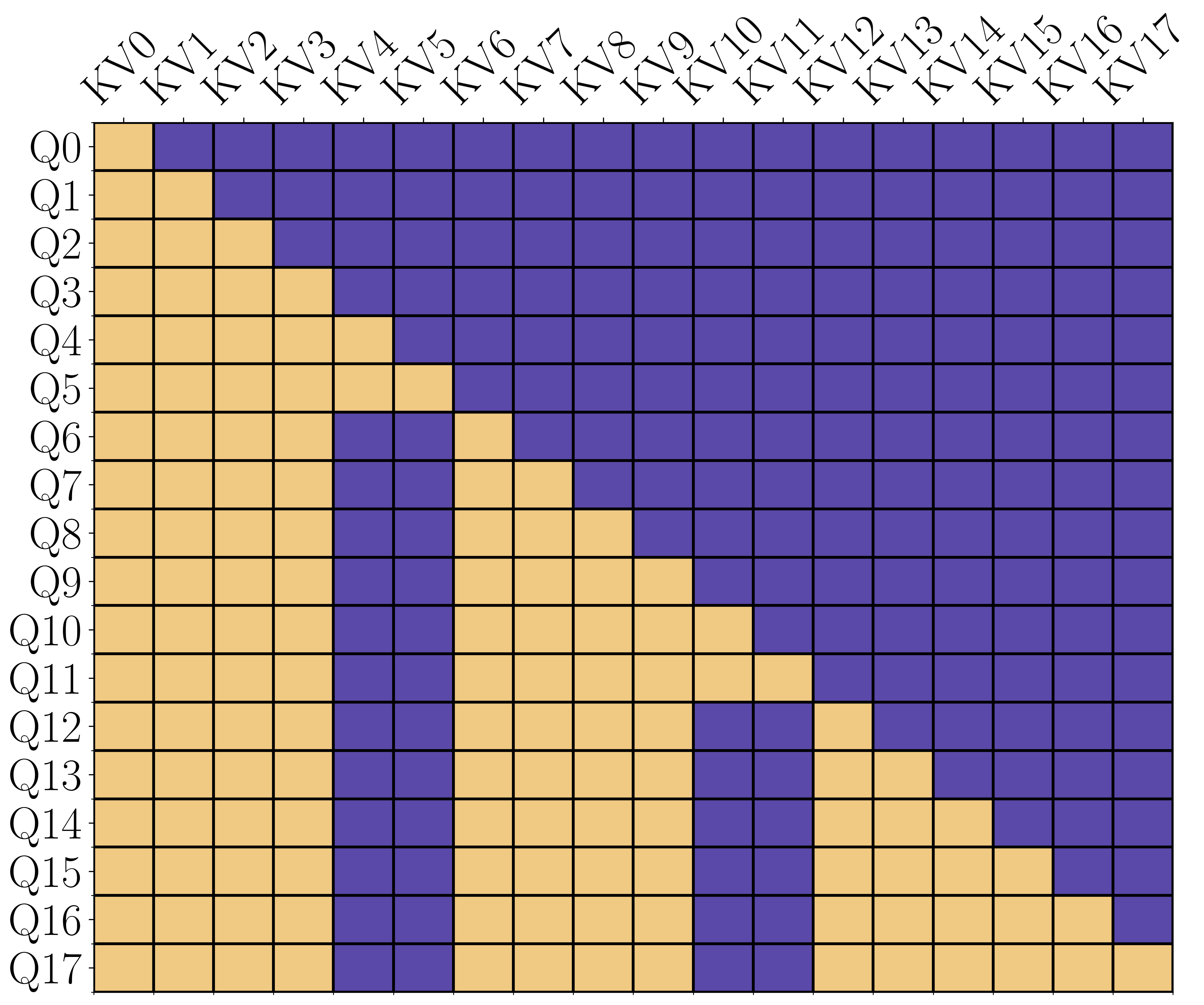}
    \caption{}
    \end{subfigure}
    \begin{subfigure}[b]{0.24\textwidth}
    \includegraphics[width=\textwidth]{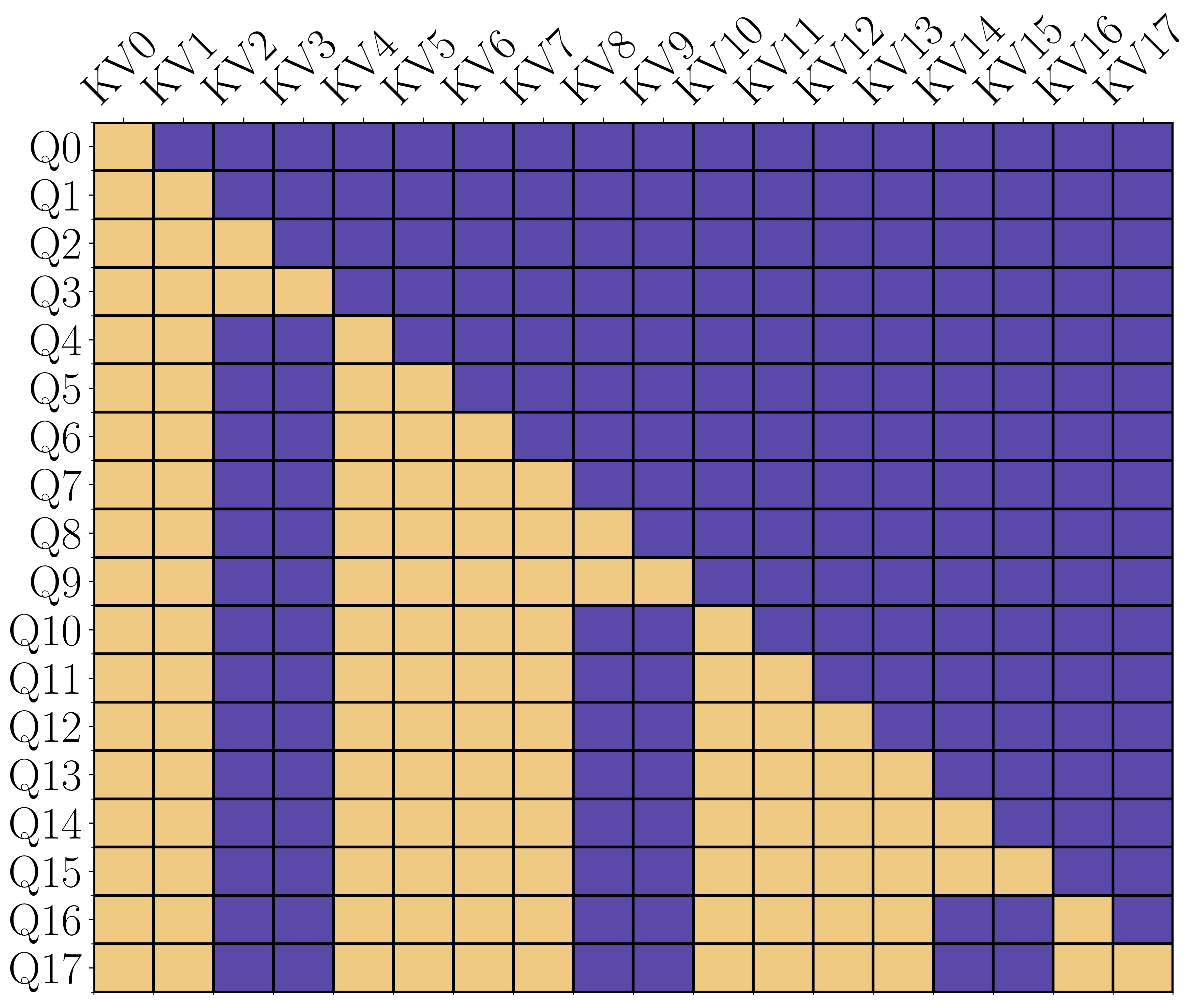}
    \caption{}
    \end{subfigure}
    \begin{subfigure}[b]{0.24\textwidth}
    \includegraphics[width=\textwidth]{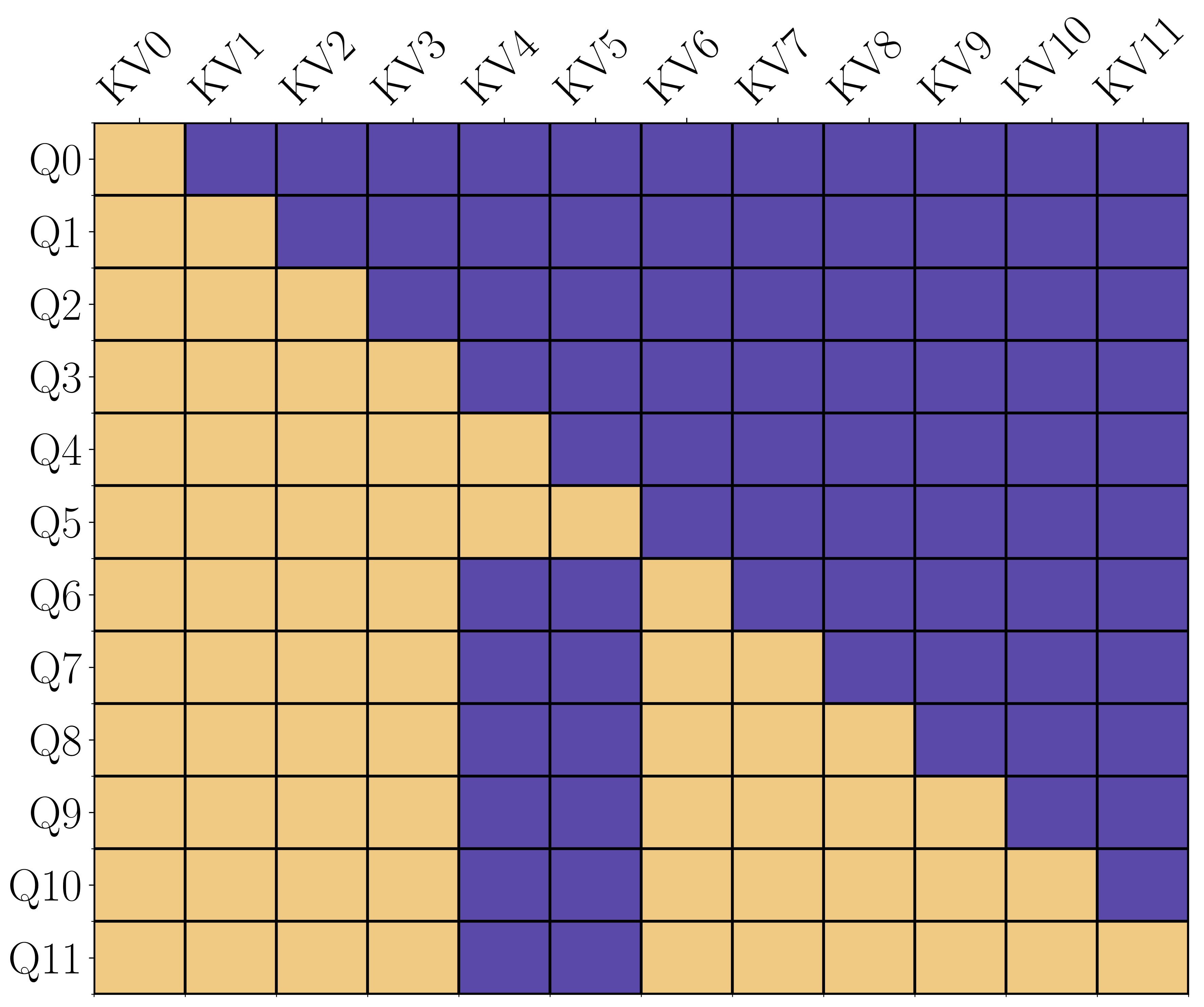}
    \caption{}
    \end{subfigure}
    \begin{subfigure}[b]{0.24\textwidth}
    \includegraphics[width=\textwidth]{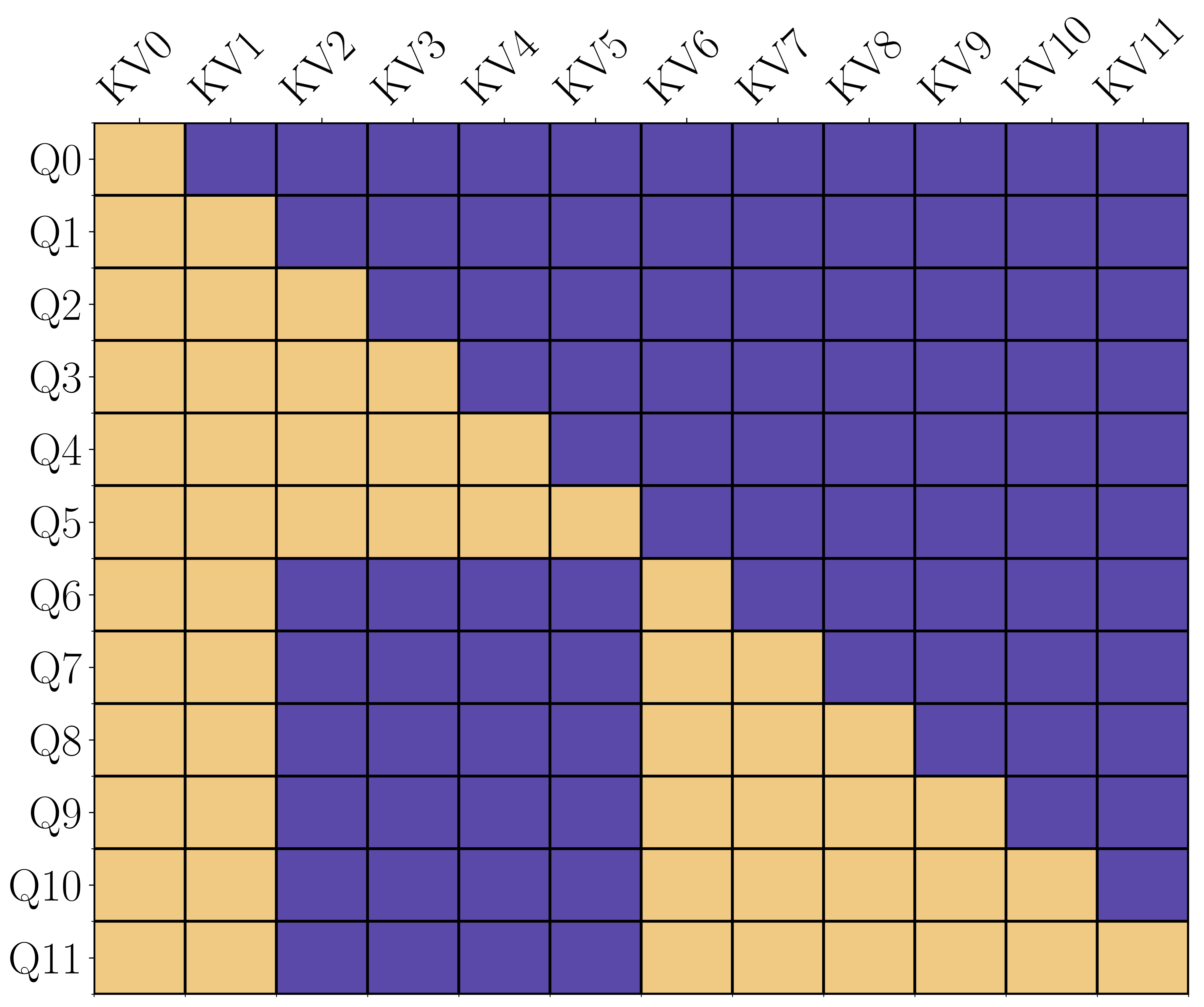}
    \caption{}
    \end{subfigure}
    \caption{Visualization of an attention masks with rolling offsets of \textbf{(a)} 0 and \textbf{(b)} -2 for a fixed $k=3$, and then with variable $k$ masking at \textbf{(c)} $k=3$ and \textbf{(d)} $k=5$ with a fixed offset of 0. \textbf{Randomized offsets and $k$ values enable supervision on MTP problems with many different prefix lengths and MTP window sizes during the same training run.}}
    \label{fig:explainer-block-mask}
\end{figure}

We visualize examples of our attention masks in \cref{fig:explainer-block-mask}. We design the mask such that for all positions corresponding to ground truth tokens outside of the MTP regions, the mask has the standard causal structure; each ground truth token attends to all upstream prefix tokens, skipping over MTP spans. Each MTP token attends to upstream ground truth tokens, in addition to upstream tokens within its own MTP span (but not other spans). 

In the masked regions where MTP is performed, the use of causal attention, rather than bidirectional or full self-attention, is not strictly necessary---whereas it \textit{is} for standard NTP training using offline cross-entropy (see \cref{app:ablations} for a discussion). However, since using a bidirectional mask in the MTP regions would represent an additional distribution shift for the pretrained student model, would also preclude the use of standard causal attention at inference time, and only provides marginal lift in our ablations (Appendix~\cref{tab:abl-l3-suprv-evals-pt1}) we choose to train our models with blocked but otherwise standard causal attention.

We also randomize the location of the MTP spans so that the model can learn to make predictions starting from any position. \cref{fig:explainer-tok-masking,fig:explainer-block-mask} illustrate how we achieve this using an offset during input preparation and mask materialization. We also experiment with randomizing the MTP span length $k$. As shown in \cref{fig:explainer-block-mask}, we parametrize our attention mask generator to allow dynamic adjustment of position, length, and number of MTP spans.

\subsection{Static vs. Adaptive Decoding Strategies}\label{sec:impl-confadapt}

Our informal analysis of the proposed objective \cref{eq:mtpo2} in the preliminaries section suggests that there may be a correlation between the confidence of the student model on its predictions and the quality of the sequence emitted. In \cref{fig:corr-analysis} we present empirical evidence of this relationship from an early experiment. In response to this confirmation we devised a simple \textit{confidence-adaptive} scheme (\textit{ConfAdapt}) that allows us to set the $k$ value dynamically at each decoding step based on this confidence heuristic. Given a threshold value $\tau$, such as $90\%$, the CA strategy finds the greatest index $k' \in (1,k_{max})$ such that $\operatorname{argmax}_j(\ell_{ij}) > \tau, \forall i \in (1,k')$, and uses that index as the $k$ value for this decoding step. This creates a dynamic range of $k$ values during a single generation which we compute statistics over to report average ``acceleration factors''.

While in the main experimental configuration we do randomize the value of $k$ during each training step, we do not directly train for any sort of dynamic decoding procedure where $k$ varies at each generation step within a single response. However, to our surprise, the MTP models we train turn out to be amenable to the simple dynamic decoding strategy described above, and it achieves pareto-optimal results when compared to static strategies.

\section{Experiments}\label{sec:exps}

We adapt pretrained NTP LMs into MTP LMs and evaluate their performance on both mathematical reasoning benchmarks and open ended knowledge intensive generative tasks.

\subsection{Setup}\label{sec:exps-setup}

We describe all aspects of our training and evaluation setup in detail in \cref{app:impl-exp-details}, but to support interpretation of the main experimental results, we succinctly note the most critical design parameters here. The pair of initial checkpoints used are Llama-3.1-8B-MagpieAlign-SFT-v0.1 (``L3.1-8B-Magpie'' hereafter, \citet{xu2024magpie}) and Qwen3-4B-Instruct-2507 (``Q3-4B-Inst-2507'' hereafter, \citet{yang2025qwen3}). In order to target a benchmark involving reasoning traces where inference efficiency is particularly useful in practice, we tune our MTP LMs on a dataset of synthetic grade school math (GSM) problems called MetaMathQA \citep{yu2023metamath}.

Based on the results of preliminary experiments, unless stated otherwise, we train with randomized block mask offsets and randomized $k$ values in the range $[2,16]$; this implicitly sets a natural choice of $k_{max}$ at 16 during inference using the \textit{ConftAdapt} scheme. We also use the ``hard teacher'' variation on \cref{eq:mtpo2} where the readout function for the teacher NTP model is the argmax rather than softmax as early experiments suggested that this lowers the student model's output entropy more rapidly. Under $k$ value randomization the expected $k$ value is closer to $9$. Over the course of the $\sim 100$k steps of training our models see a total of $\sim500$M supervised tokens (discussion and calculations in \cref{app:impl-exp-details}).

To assess the performance of our main models we select the following set of \textit{generative} benchmarks to test the technical and long form generation capabilities of our MTP LMs: GSM8K, AIME25, GPQA, BBH, IFEVAL, and CNN DailyMail. We describe the exact evaluation parameters for all datasets in \cref{app:impl-exp-details}. However, since we find the results on GSM8K to be both the strongest and also indicative of all major trends, we focus on that task in the main body of this paper.

\textbf{GSM8K CoT Fewshot} is evaluated with an additional ``Lets think step by step'' suffix to the prompt, with output truncation after emission of \verb|``Q:''| or any of the model's stop tokens. We report the ``flexible-extract'' accuracy metric and include its Std. Error in tables.

\begin{figure}[t!]
    \includegraphics[width=0.441\textwidth, trim=0cm 0cm 2.1cm 0cm, clip]{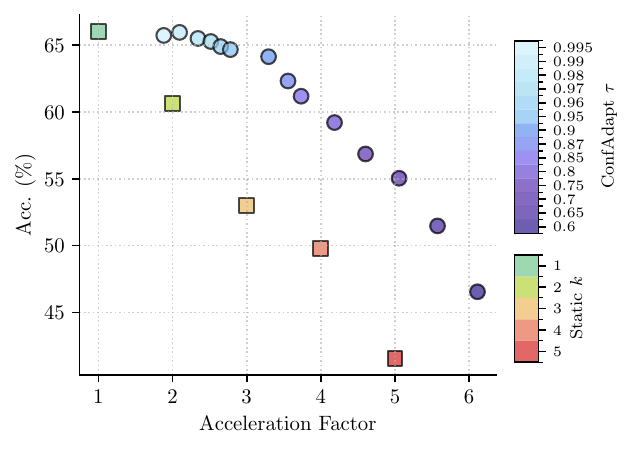}
    \includegraphics[width=0.548\textwidth, trim=0cm 0cm 0cm 0cm, clip]{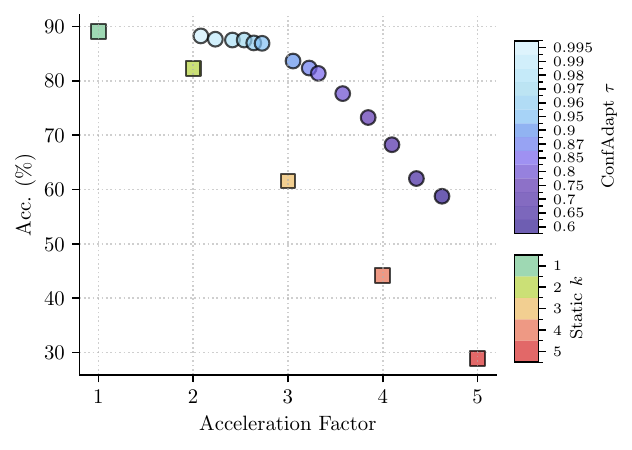}
    \caption{The performance of our (\textbf{Left}) L3.1-8B-Magpie based MTP LM and (\textbf{Right}) Q3-4B-Inst-2507 MTP LM evaluated on the GSM8K benchmark after $\sim$100k steps of training. Performance tradeoff is visualized by plotting the effective $k$ value or ``Acceleration Factor'' versus the Accuracy on the benchmark. More detailed plots showing accuracy and acceleration as a function of training steps for both models are provided in \cref{fig:dynamics-flagship-l3-gsm,fig:dynamics-flagship-q3-gsm}. \textbf{We observe that the adaptive decoding strategies achieve pareto-optimal tradeoffs between generation speed and response quality for both models.}}
    \label{fig:dynamics-flagship-l3-q3-gsm-scatter}
\end{figure}

\subsection{Accuracy vs. Acceleration}\label{sec:exps-perf-models}

In \cref{fig:dynamics-flagship-l3-q3-gsm-scatter} we summarize the performance of our L3.1-8B-Magpie based and Q3-4B-Inst-2507 based MTP models after training on MetaMathQA. Results under the \textit{ConfAdapt} scheme are visualized on the same axes as the \textit{Static} scheme by computing the average $k$ value that was chosen by the heuristic across all decoding steps over all the generations in the evaluation and include its Std. Error in tables. 

Under the \textit{Static} scheme, as $k$ increases the acceleration factor grows and the accuracy decreases. Similarly, under the \textit{ConfAdapt} scheme, as the confidence threshold $\tau$ drops, the acceleration factor increases while accuracy decays, albeit more slowly. When compared against \textit{Static} $k=1$ decoding, using the \textit{ConfAdapt} scheme at threshold of $90\%$ yields an acceleration factor of more than $3\times$ while decreasing accuracy for the L3.1-8B-Magpie by less than $3\%$; the Q3-4B-Inst-2507 achieves $3\times$ speeds at a slightly higher cost of a $7\%$ drop in accuracy. At more aggressive settings using more permissive confidence thresholds, the acceleration factors are as high as $5\times$ but cause a more significant reduction in accuracy.

\subsection{Performance Across Tasks}\label{sec:exps-perf-tasks} 

In Appendix~\cref{tab:math-evals,tab:general-evals}, we present the same pair of models but evaluated over a larger set of 6 evaluation tasks. We observe that even though both models were trained on only MetaMathQA, transfer learning occurs. As the two initial models have different strengths before MTP adaptation, both at step 0 and after training is complete, their performance under \textit{Static} $k=1$ decoding varies considerably. However, we generally observe that larger average acceleration factors occur on the GSM8K and BBH benchmarks, and that for the more open-ended generative tasks like CNN DailyMail, we see lower acceleration factors across all decoding strategies.

\subsection{Ablations on the MTP Objective}\label{sec:exps-abls}

In addition to the two main models that we showcase above, in the Appendix we include results for ablations where we change the training dataset and where we modify the various parameters of our MTP training objective. These ablations are presented in \cref{app:ablations} via \cref{tab:abl-l3-suprv-evals-pt1,tab:abl-l3-suprv-evals-pt2} but we can summarize those results as supportive of our choice of hyperparameters and objective configuration for the main models. To highlight just a few of the ablated settings, we observe the best overall performance when we utilize the hard teacher distribution under the argmax readout as targets, when we randomize the $k$ values during training, when we use causal masking across the MTP regions, and when there is no auxiliary loss term for NTP on prefix tokens.

\begin{wrapfigure}{r}{0.5\textwidth}
    \vspace{-0.5cm}
    \includegraphics[width=\linewidth]{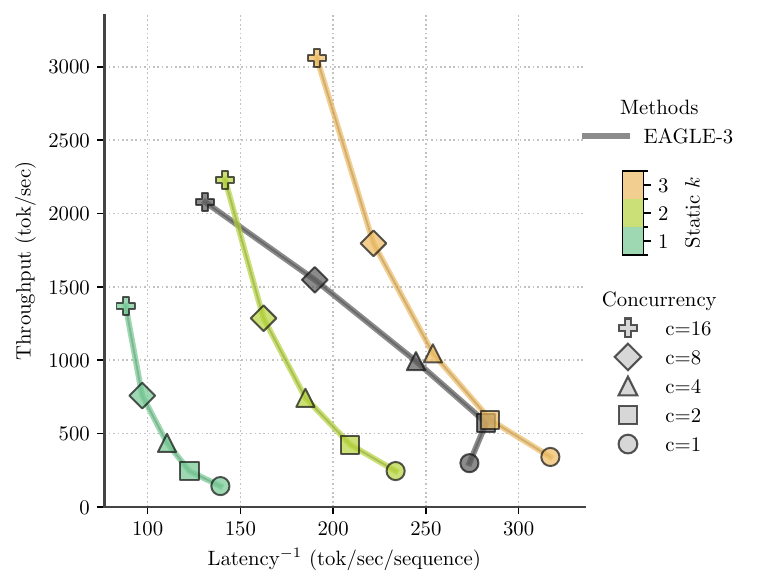}
    \caption{Throughput vs. latency tradeoff for L3.1-8B-Magpie. Throughput is the total tokens per second emitted by the server, latency is the tokens per second per request, and concurrency is the number of parallel requests made to the server. \textbf{MTP decoding smoothly trades latency for throughput and is competitive with EAGLE-3 under our static k=3 strategy.}}
    \label{fig:tpt-pareto-8b-reduced}
    \vspace{0.2cm}
\end{wrapfigure}

\subsection{Throughput vs. Latency}\label{sec:tpt-vs-latency}

Finally, while the ``Acceleration Factor'' metric reported above is an interpretable measure of speedup for a MTP approach, real world gains may be different. Therefore, we also analyze our approach using an integration with the inference framework SGLang \citep{zheng2024sglang}; their engine supports caching, batching, and model parallelism features and enables a comparison to baselines like EAGLE-3 speculative decoding \citep{li2025eagle}. Visualized as a pareto frontier between throughput and latency, in \cref{fig:tpt-pareto-8b-reduced} we observe that our approach is competitive with EAGLE-3 under our static k=3 strategy. However, in a companion chart, Appendix~\cref{fig:tpt-pareto-8b-32b-full}, we also see that the ConfAdapt scheme's overhead limits the throughput advantage concurrency can provide in that setting. Note that the irregularity for EAGLE-3 between $c=1$ and $c=2$ is likely caused by a change in engine behavior behind the scenes when switching between single prompt and batching. See \cref{app:tpt-vs-latency} for more in depth discussion of these implementation details and results across more strategies and a larger 32B parameter model.

\subsection{Limitations}\label{sec:limitations}

While we present a series of ablations to demonstrate the efficacy of our proposed objective, early experiments and the outcomes of certain ablations suggest that there are many avenues for improvement. For example, while we observe non-trivial lift when using randomized $k$-values rather than static ones during training, there are other ways we could have performed that ablation that might be more compelling. We also explored utilizing a non-static curriculum over $k$ values but those experiments were inconclusive. We considered directly penalizing the entropy of the student model to collapse its distribution faster, but we expect that other approaches such as a second stage of traditional online RL with complete multi-step roll outs might prove more effective than the penalty term approach. Finally, the pair of datasets and limited training budgets we utilize yield surprisingly strong results but we expect that scaling up the recipe (dataset, model size, training compute) could all help produce more performant versions of our models. We will release our training code and model checkpoints upon publication so that others can explore some of these directions in future research; \cref{app:future-work} discusses a few of these avenues in more detail.

\section{Conclusion}\label{sec:conc}

Our methodology produces MTP LMs with accelerated decoding abilities using a straightforward recipe and modest computational resources. While techniques in the speculative decoding literature achieve similar levels of acceleration under certain conditions, they require the training of additional speculator modules and the development and maintenance of complex inference implementations to handle the proposal and verification steps. Our approach is \textit{much} simpler but achieves similar results. More importantly, our results suggest that decoding acceleration need not rely solely on inference-time mechanisms. By absorbing some modeling complexity into the main parameters themselves, our MTP strategy provides a complementary and largely orthogonal axis to existing accelerated decoding methods.




\section*{Acknowledgements}

This work was supported by DARPA TIAMAT, the NSF TRAILS Institute (2229885), and Coefficient Giving. Computing resources for this project were supported in part by the Swiss National Supercomputing Center (CSCS).  Prepared in collaboration with Lawrence Livermore National Lab (LLNL) under Contract DE-AC52-07NA27344 and supported by the LLNL-LDRD Program under Project No. 24-ERD-010 (LLNL-CONF-2015543).

We would also like to acknowledge Jonas Geiping and Sean McLeish for helpful discussions throughout the research process on infrastructure and tooling for experimentation on the public supercomputers at CSCS and LLNL.

\section*{Responsible Disclosure of LLM Use}

Following ``Policy 1'' we briefly describe the use of LLMs and coding agents in our research. Throughout the implementation process, single functions and standalone scripts were generated and iterated upon with the help of LLMs. Throughout the experimentation process, LLMs were used to help write data transformation and plotting code to analyze results. Finally, the integration of our method with the production ready SGLang inference engine was primarily implemented by a coding agent under the guidance of one of the authors; the author debugged and sanity checked the model's work throughout the integration process.

\bibliography{references}
\bibliographystyle{colm2026_conference}

\newpage
\appendix
\crefalias{section}{appendix}

\section{Directions for Future Research.}\label{app:future-work}
We include a discussion of a number of promising avenues for future work that can build on top of our results and model checkpoints:

\textbf{(i)} Reducing the entropy of our MTP LM to zero is the goal of the training process, but our training process may not be the best procedure; future work could explore other strategies for decreasing output entropy. We expect that as entropy in the output distribution approaches zero, we can conceivably produce many more tokens during each forward pass while maintaining coherence.

\textbf{(ii)} Even under accelerated decoding, our goal is to still faithfully model the highly entropic distribution of natural text, which is in tension with our acceleration scheme's implicit low entropy requirements. Some simple decoding strategies that would force non-determinism purely at test time are possible (e.g. sampling from the softmax whenever $k=1$), but future work could incorporate a source of randomness more naively into the model to increase roll out diversity.

\textbf{(iii)} Our models could be used for self-speculative decoding. In this work we specifically avoid incorporating a verification pass to explore whether or not fast generation of $k$ tokens in a single shot without verification is even feasible. However, in principle, the proposals made by our MTP model could be used for self-speculative decoding. The MTP model could verify its own predictions during each subsequent forward pass to achieve more modest but lossless speedups. We expect that our KV caching strategy is amenable to an additional step where previously emitted tokens are verified before committing them to the cache (see \cref{sec:impl-kvcache}).

\textbf{(iv)} Our simple MTP implementation involves using a trainable mask token to extend the sequence passed into the model by $k$ positions, creating additional slots for computation within the model, but other MTP architectures are possible. In principle, adding small auxiliary prediction heads or modules can enable MTP and it is possible that these less computationally intensive implementations are as readily trainable and performant as our proposal, even under the same objective.

\textbf{(v)} Our online distillation objective can be viewed as a simple type of RL, but a full-fledged policy optimization approach may yield better results. For simplicity and parallelizability, during training we use a single forward pass to produce just one block of $k$ tokens at each masked region within a training sample. However, a more long horizon approach is possible where the student MTP model rolls out its own full sequence comprising many chunks of $k$ tokens over multiple forward passes that is scored using a more traditional reward signal. This could result in a beneficial collapse in entropy that improves output quality and acceleration.

\textbf{(vi)} We present a preliminary set of results on how our approach would behave in a production inference environment via a prototype implementation within SGLang (see \cref{app:tpt-vs-latency} for details). While we observe promising scaling across the board under the static $k$ strategy and strong results for the ConfAdapt scheme on single user requests, our naive implementation and the overheads associated with the ConfAdapt scheme limit the scaling we observe under continuous batching. As we are not systems experts, we expect that there is room for deeper engineering optimizations that would address these issues in production.

\newpage
\section{Extended Implementation and Experimental Details}\label{app:impl-exp-details}

\begin{algorithm}[h!]
\caption{Formal definition of the training step for our online MTP objective with randomized k in $[2,16]$ and randomized block mask offsets in the ``hard teacher'' setting. Note that teacher is always evaluated without computing its gradients.\\\\
\textsc{PrepareMTPBatch} materializes the masked inputs and blocked attention pattern illustrated in Figs.~\ref{fig:explainer-tok-masking} and \ref{fig:explainer-block-mask}: mask $\mathcal{P}$ selects predicted regions, mask $\mathcal{M}$ denotes \texttt{<MTP>} masks, and $\mathcal{B}$ is the block attention mask used by both the student and teacher models.\\
\textsc{InsertStudentTokens} inserts the leading $k-1$ tokens predicted by the student into prepared sequence at the corresponding masked positions.\\
\textsc{OptimizerStep} refers to the AdamW update in our experiments, but any gradient-based optimizer may be used.}
\label{alg:singleshot-online-train-step}
\begin{algorithmic}[1]
\REQUIRE Student parameters $\theta_S$, frozen teacher parameters $\theta_T$, batch of raw token sequences $X$.
\STATE Sample $k \sim \operatorname{Unif}\{2,\dots,16\}$ and a rolling offset $o$.
\STATE $(\widetilde{X}, \mathcal{P}, \mathcal{M}, \mathcal{B}) \leftarrow \textsc{PrepareMTPBatch}(X, k, o)$
\STATE $L^S \leftarrow f_{\theta_S}(\widetilde{X},\mathcal{B})$
\STATE $Z^S \leftarrow L^S[\mathcal{P}] \in \mathbb{R}^{R \times k \times |\mathcal{V}|}$ \COMMENT{$R$: active MTP regions in the batch}
\STATE $\widehat{Y} \leftarrow \arg\max Z^S \in \mathcal{V}^{R \times k}$
\STATE $\widetilde{X}^{\,\mathrm{sf}} \leftarrow \textsc{InsertStudentTokens}(\widetilde{X}, \mathcal{M}, \widehat{Y}_{:,1:k-1})$
\STATE $L^T \leftarrow f_{\theta_T}(\widetilde{X}^{\,\mathrm{sf}},\mathcal{B})$
\STATE $Z^T \leftarrow L^T[\mathcal{P}] \in \mathbb{R}^{R \times k \times |\mathcal{V}|}$
\STATE $\overline{Y} \leftarrow \arg\max Z^T \in \mathcal{V}^{R \times k}$ \COMMENT{``hard teacher'' labels.}
\STATE For each region $r$ with prefix $x_{1:i_r}$, let $\hat{y}_r := \widehat{Y}_r$ and define
\STATE $\overline{Y}_{r,j} = \arg\max P_{\theta_T}(\cdot \mid \hat{y}_{r,1:j-1} \oplus x_{1:i_r})$ for $j=1,\dots,k$
\STATE $P_{\theta_T}^{\mathrm{hard}}(y_{1:k} \mid x_{1:i_r}, \hat{y}_r) := \mathbf{1}\!\left[y_{1:k} = \overline{Y}_r\right]$
\STATE $\mathcal{L}_{MTP} = - \dfrac{1}{R}\sum_{r=1}^{R} P^{hard}_{\theta_T}(y' | x_{1:i_{r}}) \log P_{\theta_S}(y' | x_{1:i_{r}})$
\STATE $\theta_S \leftarrow \textsc{OptimizerStep}\!\left(\theta_S, \nabla_{\theta_S}\mathcal{L}_{MTP}\right)$
\end{algorithmic}
\end{algorithm}

\paragraph{Initialization}

To minimize training requirements, we initialize the student MTP policy and teacher NTP model from the same set of strong pretrained weights from a NTP model. During training, we keep the teacher's parameters frozen while allowing all parameters in the MTP model to freely train.  We randomly initialize a special \verb|<MTP>| token (as shown in \cref{fig:explainer-tok-masking}) in the embedding and un-embedding matrices of the MTP LM according to the mean and variance of the existing pretrained embeddings.

\paragraph{KV Cache Management}\label{sec:impl-kvcache}
 
During inference, our MTP LM appends mask tokens to the prefix before the forward pass. These tokens should be removed from the KV cache after they are used for prediction.  In our implementation, the first prediction pass on a sequence generates $k$ KV pairs, one for each token predicted.  After this pass is complete, the mask tokens are removed from the prefix and the corresponding KV values are popped off the top of the cache. Then, $2k-1$ tokens are appended to the prefix; the $k$ newly generated tokens and $k-1$ new mask tokens for the next forward pass.  We proceed in this way, popping stale values off the cache and adding the new tokens and masks before every forward pass. This scheme allows us to save KV cache space while also using a standard causal mask at inference time (rather than a blocked mask like in~\cref{fig:explainer-block-mask}) which is typically more efficient in most transformer implementations.

\paragraph{Stopping criteria.}

Similar to KVCaching, the natural generalization of stop-token handling, eg. EOS tokens, from the NTP setting to our MTP model is straightforward but still requires explanation. 

During training, since we primarily consider instruction following datasets with variable lengths, we have to choose how to train the model to emit an EOS token. Our convention is to treat the first MTP region in which an EOS appears in the ground truth sequence as the determiner of the last MTP region for this sample in the training data. Then, we pad out the sequence with more EOS tokens to fill this last $k$ token region. Beyond this region, the rest of the context window is filled with padding and ignored completely by the computation and objective. 

At inference time, we therefore hope that the model has then learned to emit at least one EOS token in the appropriate point in the sequence. During generation, if any of the tokens emitted by the MTP model in a given $k$ token decoding step are stop-tokens, we simply call this the final step of generation and terminate, trimming off any tokens predicted after the first EOS. We find that this simple setup provides the MTP model the ability to reliably generate long sequences that truncate at logical positions that are also not exact multiples of $k$.

\paragraph{Pretrained models.}

\begin{figure}[h!]
    \centering
    \includegraphics[width=0.6\textwidth, trim=0cm 0cm 0cm 0cm, clip]{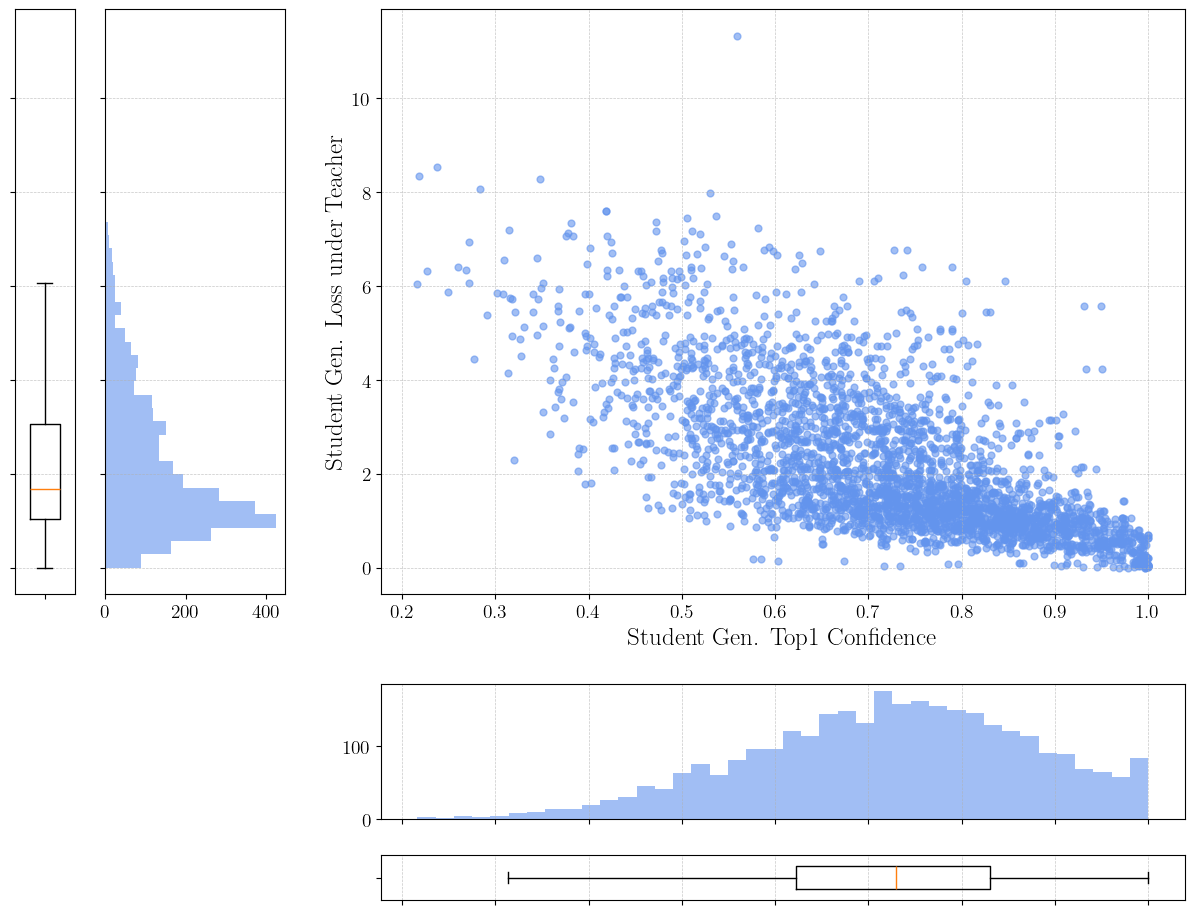}
    \caption{The correlation between student confidence and generation quality for a preliminary experiment where we train a Llama-3.1-8B Base model on generic webtext under our MTP objective. We evaluate the quality of its completions to prefixes in a validation set (loss under the teacher model, lower is better) and correlate this against the student's confidence in its own generations (avg. probability assigned to top token in all $k$ positions). \textbf{The correlation between confidence and quality allows us to design a simple adaptive decoding strategy that limits the number of tokens sampled if the model is not sufficiently confident at any point during the generation process.}}
    \label{fig:corr-analysis}
\end{figure}

Initially we prototyped our approach using the Llama-3.2-1B base checkpoint \citep{grattafiori2024llama} to enable low cost, rapid iteration of our design. Then we upgraded to the Llama-3.1-8B base checkpoint \citep{grattafiori2024llama} to improve overall performance while exploring loss variations and training hyperparameters (\cref{fig:corr-analysis} is derived from experiments with that model). Finally, based on intuitions that post-trained models generally have lower entropy in their NTP distribution due to training on narrow domains using objectives and datasets that empirically reduce a model's entropy, we switched to post-trained initial checkpoints for the final set of experiments and ablations. The pair of post-trained initial checkpoints used are Llama-3.1-8B-MagpieAlign-SFT-v0.1 (``L3.1-8B-Magpie'' throughout, \citet{xu2024magpie}) and Qwen3-4B-Instruct-2507 (``Q3-4B-Inst-2507'' throughout, \citet{yang2025qwen3}).

\paragraph{Finetuning data.}

In order to target benchmarks involving reasoning traces where inference efficiency is most useful in practice, we train our MTP LMs on a dataset of synthetic grade school math (GSM) problems called MetaMathQA \citep{yu2023metamath}. The data is a synthetically generated series of GSM problems with reasoning traces and answers built by taking an existing set of previously released, hand-written questions, and augmenting them using a large API based model. We note that this dataset is based on the train split of the official GSM8K dataset only, the official test set questions used for benchmarking are not included or used as seed questions. For our experimentation, we also split the 395k raw rows into subsets for training and validation that are stratified by the unique set of seed questions such that augmentations of a single seed question fall into the train set, or the validation set, but not both.

In an ablation, we also consider a more general set of finetuning data: Magpie-Pro-300k-v0.1 and Magpie-Reasoning-150k (``Magpie'' hereafter, \citet{xu2024magpie}). These are also synthetically generated datasets covering a wide range of instruction following and reasoning tasks, but we specifically choose it due to the release of a publicly released model checkpoint that was post-trained directly on these datasets (the aforementioned L3.1-8B-Magpie). We hypothesize that the affinity between the pretrained initialization of the student policy and its ability to rapidly learn the MTP behavior may be linked and so we compare the impact of this training data choice, holding all other settings constant.

\paragraph{Chat Templating.}

Both of the main pretrained models we use come equipped with a chat template. However, preliminary experiments indicated that it was difficult to ensure that our training data and evaluation data was prepared in such a way that the model responded accurately under standard $k=1$ decoding at the end of training. As a result, we experimented with various combinations before landing on the following configuration. For the L3.1-8B-Magpie model, we apply the chat template included with its tokenizer for all training and evaluation datasets as this performed well. However, for Q3-4B-Inst-2507, after experimenting with its chat template and observing poor results, we settled on preparing the training and evaluation data by joining ``input'' and ``response'' pairs using just a simple ``\verb|\n\n|'' string and prepending a BOS token to every full input. In all tables, the setting for the model is noted under its name.

\paragraph{Hyperparameters.}

In addition to the initialization and training datasets, we adopt a standard set of hyperparameters for elements of the training pipeline not specific to our MTP implementation and objective. We train our models using the AdamW optimizer using PyTorch default settings and a learning rate schedule of 2000 steps of warm-up followed by a constant peak learning rate of $1e-5$; our goal is to keep the model in a ``finetuning'' regime where the student's initial capabilities and behaviors are maintained whilst the MTP abilities are acquired. All parameters in the student model are made trainable, while the teacher model remains frozen.

As our primary training dataset, MetaMathQA, contains relatively short samples (avg. $\sim 220$ tokens under the Llama 3 tokenizer), we cap the overall sequence length (training context window, $N$) at 160 tokens for all MetaMathQA runs. For ablations using the Magpie datasets whose average lengths are longer, we use a sequence length of 1024 tokens. For experiments with MetaMathQA, we use a global batch size of 128 sequences across all devices training in data parallel, and for the Magpie dataset experiments we use a global batch size of 16. While not perfectly equivalent, the number of tokens per optimization step is therefore within the same order of magnitude for both experiments. Based on a desire to balance the supervision density of our objective with the length of prefixes that the model is exposed to, we set the number of MTP regions $M$ in a given sequence during training to be $M=N/(2k)$ as illustrated in \cref{fig:explainer-tok-masking} (we use the largest $k$ value 16 as the value when choosing $M$, eg. $M=N/(2k_{max})$).

For implementation simplicity, while the offset and $k$ value are randomized during training, $M$ is left fixed meaning that all sampled values of offset and $k$ produce a lower number of supervised positions (eg. the sum of the ``Pred:'' row in \cref{fig:explainer-tok-masking}) than the maximum possible which is achieved when the offset is 0 and the sampled $k$ is the maximum. We use a batch size of 128 sequences for the models trained on MetaMathQA, and a batch size of 16 for the models trained on the Magpie datasets. While not scaled exactly, due to the $> 5\times$ larger sequence length for the latter, the reduced batch size still achieves in the same order of magnitude number of tokens per optimization step. Over the course of approximately 100k training iterations, for the MetaMathQA models based on the setting of $M=N/(2k_{max})$, this equates to an upper bound of 1B supervised tokens if assuming a fixed value $k=k_{max}$ and an offset of 0. Since $160 /(2*16) = 5$ regions, we have $100e3 * 128 * (5 * 16) = 1.024$B tokens. For the actual runs under randomization, in expectation the $k$ value is closer to half of the max, so our models only see a total of 500M supervised tokens. 

As illustrated in \cref{fig:explainer-tok-masking}, more tokens of raw input are required as input at each step than are actually supervised if $k$ is randomized following the static $M$ setting described above. As there are about 50M tokens available in the MetaMathQA dataset assuming the Llama 3 tokenizer and a truncation length of 160, this results in $\sim35$ epochs of training for the L3.1-8B-Magpie based model. While early experiments indicate that overfitting is possible at more extreme durations, due to the offset and $k$ value randomization strategies, the number of unique batches of prefix and MTP prediction problems in a dataset with $D$ rows is actually $>D$. These additional views of the same sample---approximately proportional to the unique combinations of block offset and $k$ possible under a given set of hyperparameters---likely make the effective number of epochs on the same exact batch of prepared and masked inputs much lower than (steps * microbatch size)/$D$, if not close to even just 1. We believe that the result of transfer learning from MetaMathQA to benchmarks beyond GSM8K also suggest that overfitting isn't necessarily an issue in our specific experimental settings.

The main models containing 8B and 4B parameters are both trainable at the settings described using one node of 4$\times$GH200 GPUs using FSDP based model and data parallelism. The training runs to 100k steps take approximately 24-36 hours of continuous runtime depending on the model. Each benchmark evaluation can be performed on a single GH200 GPU and runtime varies depending on the specific settings of $k$, the size and average response length for the benchmark itself.

\paragraph{Evaluation metrics and benchmarks.}

During early experimentation, we used our held out validation subset of MetaMathQA to track metrics like confidence and quality of the student roll outs under the teacher (as in \cref{fig:corr-analysis}). Then, for a comprehensive evaluation of our main models we selected a series of mathematical and general instruction following benchmark datasets from the Eleuther LM Evaluation Harness~\citep{eval-harness}. We specifically chose the following set of \textit{generative} benchmarks to test the technical and long form generation capabilities of our MTP LMs: GSM8K, AIME25, GPQA, BBH, IFEVAL, and CNN DailyMail. We refer the reader to the harness repo and each respective dataset's documentation for a detailed descriptions of each but we list the specific used settings in the evaluation harness below.

\textbf{GSM8K CoT Fewshot} is evaluated with an additional ``Lets think step by step'' suffix to the prompt, with output truncation after emission of \verb|``Q:''| or any of the model's stop tokens. We report the ``flexible-extract'' accuracy metric and its Std. Error.

\textbf{AIME25} is evaluated with output truncation after emission of any of the model's stop tokens. We report the ``exact-match'' accuracy metric and its Std. Error.

\textbf{GPQA Main CoT Fewshot} is evaluated with output truncation after emission of any of the model's stop tokens. We report the ``flexible-extract'' accuracy metric and its Std. Error.

\textbf{Big Bench Hard CoT Fewshot} is evaluated with output truncation after emission of \verb|``\\n\\n''|, \verb|``Q:''|, or any of the model's stop tokens. We report the exact-match accuracy metric and its Std. Error under the ``get-answer'' extraction method.

\textbf{IFEval} is evaluated with output truncation after emission of \verb|``Q:''| or any of the model's stop tokens. We report the ``Prompt-level'' exact-match accuracy metric and its Std. Error.

\textbf{CNN DailyMail Summarization} is evaluated with output truncation after emission of any of the model's stop tokens. We report the ROUGE-L performance metric and its Std. Error.

\newpage
\section{Extended Results}

\subsection{Quality vs. Confidence Correlation}\label{app:conf-qual-corr}

\subsection{Performance over the course of training.}
\begin{figure*}[h!]
    \includegraphics[width=\textwidth]{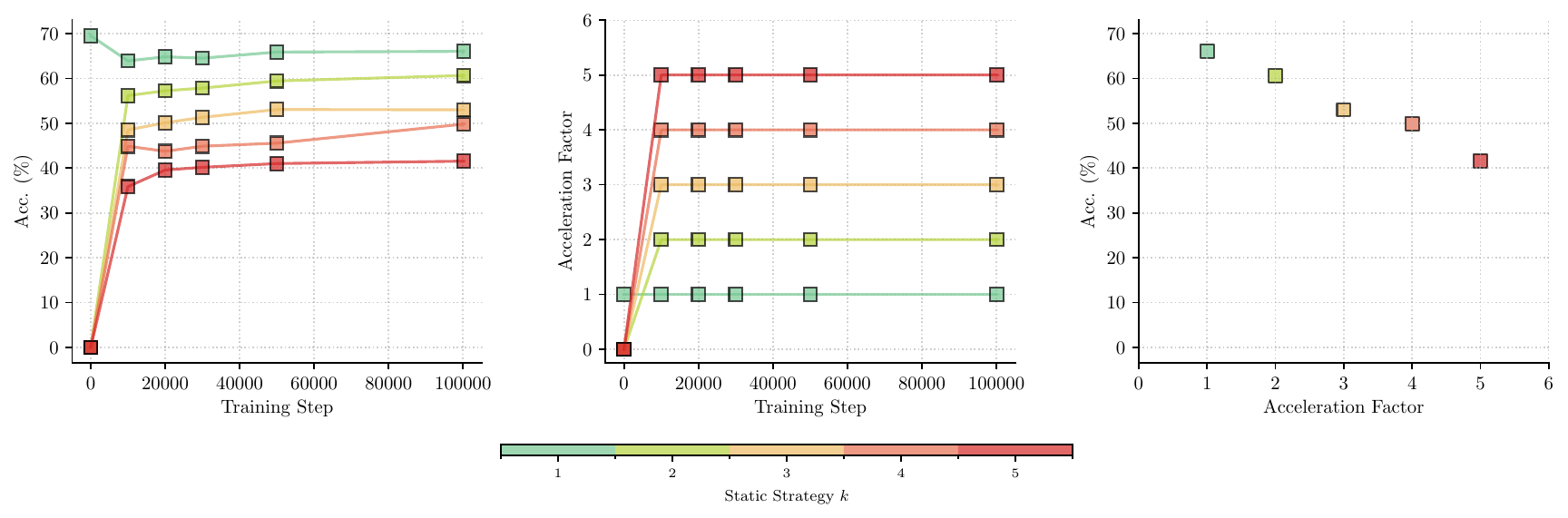}
    \includegraphics[width=\textwidth]{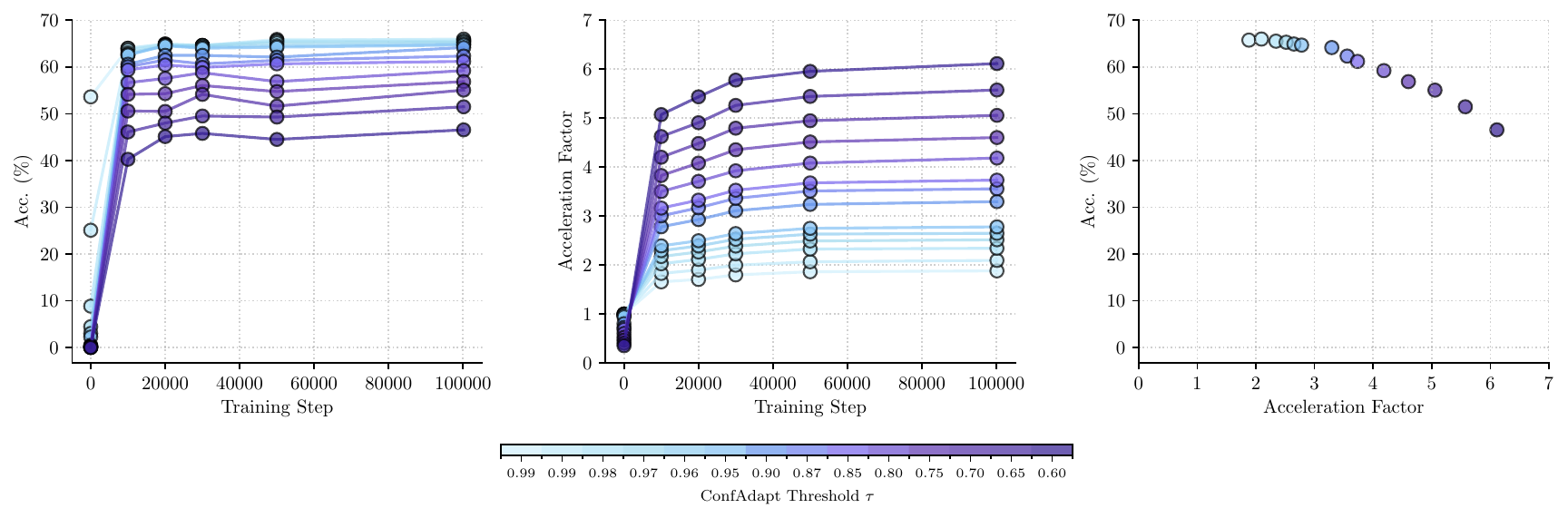}
    \caption{(Left) Accuracy and (middle) acceleration dynamics for L3.1-8B-Magpie evaluated on GSM8K over the course of training. (Right) The pareto frontier of accuracy versus acceleration based on the final model checkpoint evaluated across all decoding strategies. We find that performance and acceleration stabilize around 50k training steps, and the adaptive decoding strategies achieve pareto-optimal tradeoffs between generation speed and response quality.}
    \label{fig:dynamics-flagship-l3-gsm}
\end{figure*}

\begin{figure*}[h!]
    \includegraphics[width=\textwidth]{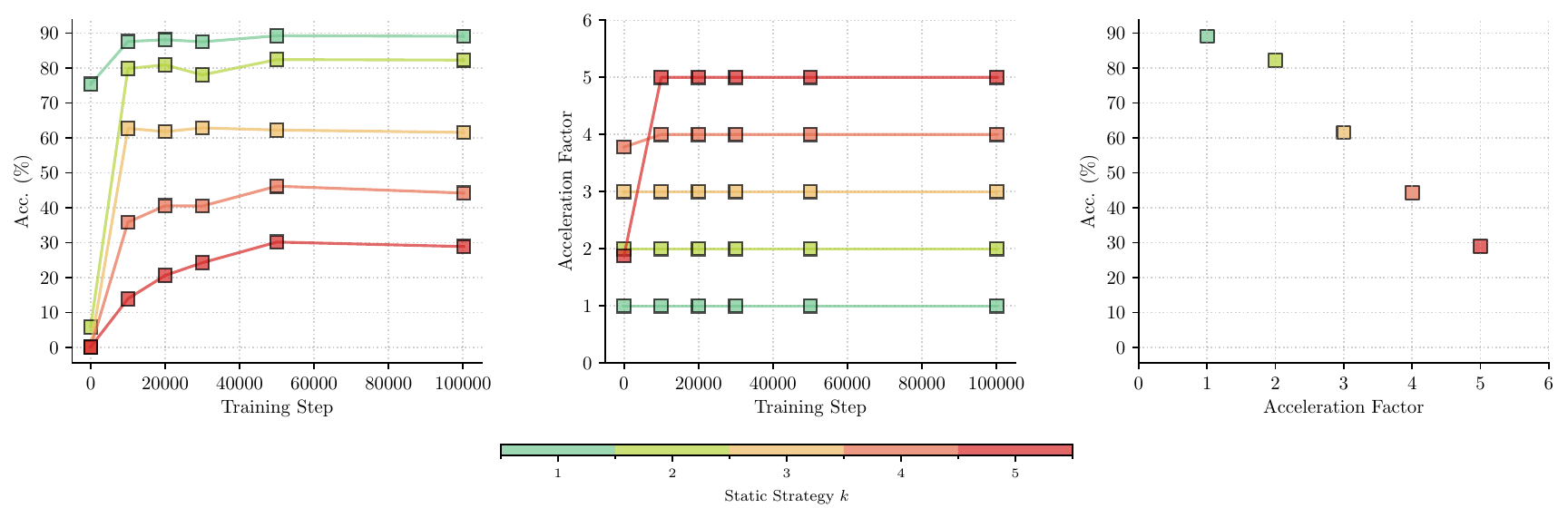}
    \includegraphics[width=\textwidth]{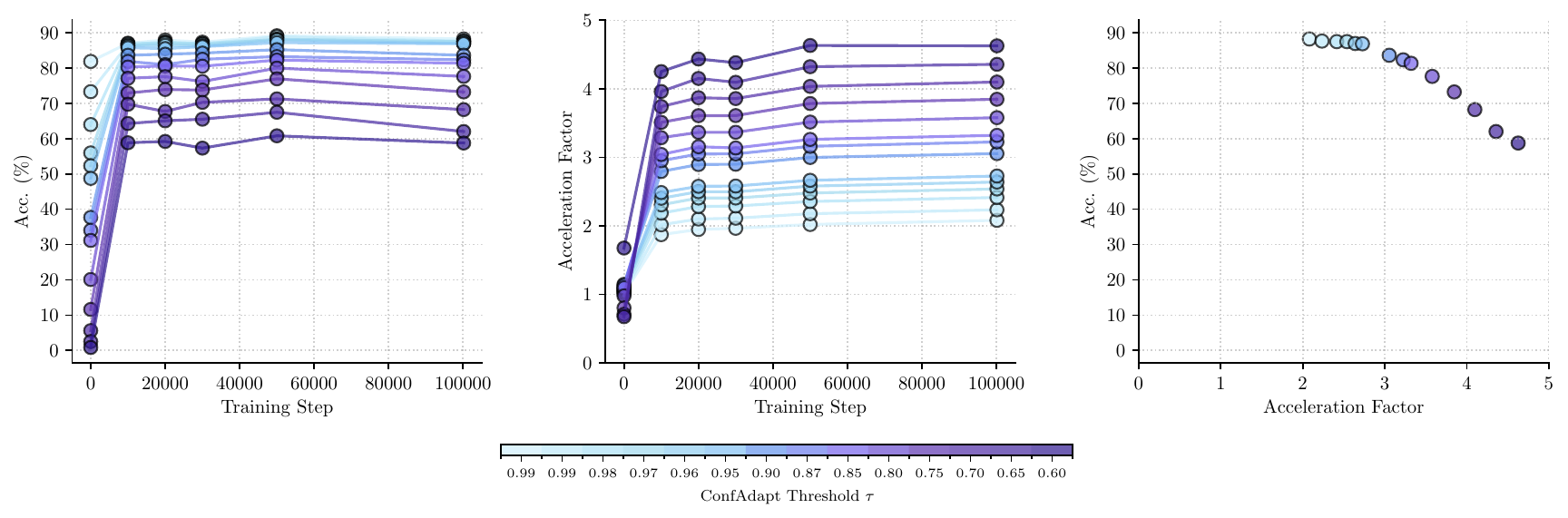}
    \caption{(Left) Accuracy and (middle) acceleration dynamics for Q3-4B-Inst-2507 evaluated on GSM8K over the course of training. (Right) The pareto frontier of accuracy versus acceleration based on the final model checkpoint evaluated across all decoding strategies. We find that performance and acceleration stabilize around 50k training steps, and the adaptive decoding strategies achieve pareto-optimal tradeoffs between generation speed and response quality.}
    \label{fig:dynamics-flagship-q3-gsm}
\end{figure*}

\begin{figure*}[t!]
    \includegraphics[width=0.435\textwidth, trim=0cm 0cm 2.1cm 0cm, clip]{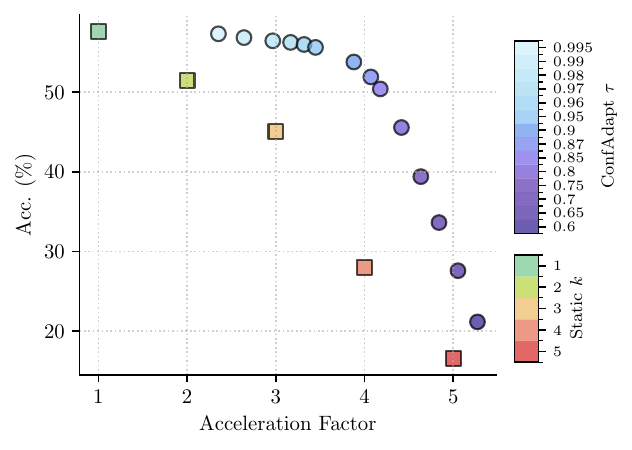}
    \includegraphics[width=0.554\textwidth, trim=0cm 0cm 0cm 0cm, clip]{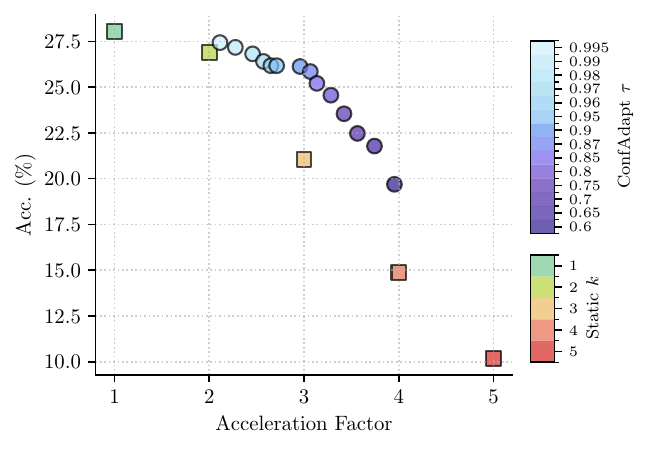}
    \caption{The performance of our (\textbf{Left}) L3.1-8B-Magpie based MTP LM and (\textbf{Right}) Q3-4B-Inst-2507 MTP LM evaluated on the BBH benchmark after $\sim$100k steps of training. Performance tradeoff is visualized by plotting the effective $k$ value or ``Acceleration Factor'' versus the Accuracy on the benchmark. More detailed plots showing accuracy and acceleration as a function of training for both models are provided in \cref{fig:dynamics-flagship-l3-bbh,fig:dynamics-flagship-q3-bbh}. We observe that the adaptive decoding strategies achieve pareto-optimal tradeoffs between generation speed and response quality for both models.}
    \label{fig:dynamics-flagship-l3-q3-bbh-scatter}
\end{figure*}

\begin{figure*}[h!]
    \includegraphics[width=\textwidth]{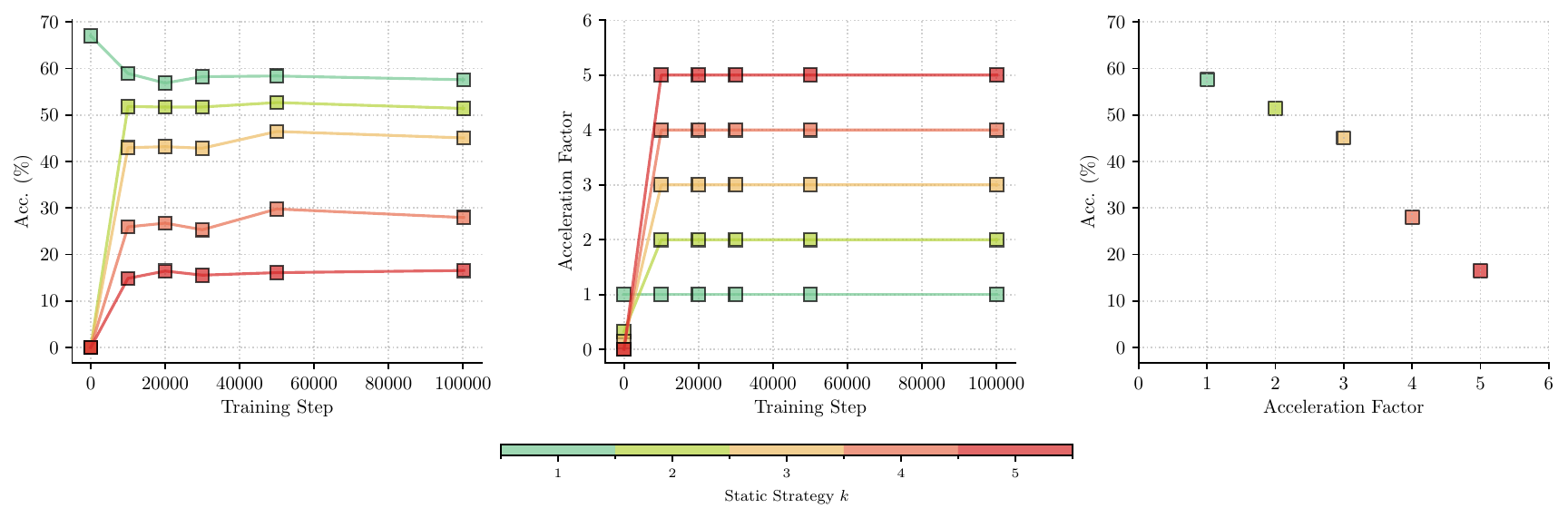}
    \includegraphics[width=\textwidth]{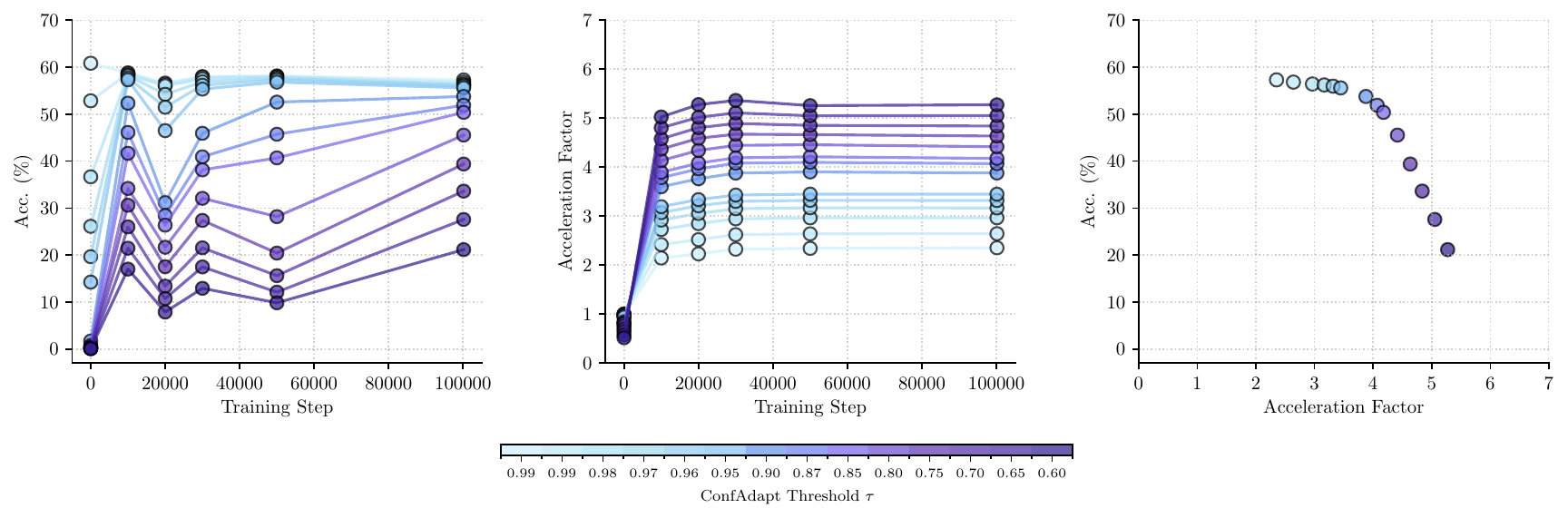}
    \caption{(Left) Accuracy and (middle) acceleration dynamics for L3.1-8B-Magpie evaluated on BBH over the course of training. (Right) The pareto frontier of accuracy versus acceleration based on the final model checkpoint evaluated across all decoding strategies. We find that performance and acceleration stabilize around 50k training steps, and the adaptive decoding strategies achieve pareto-optimal tradeoffs between generation speed and response quality.}
    \label{fig:dynamics-flagship-l3-bbh}
\end{figure*}

\begin{figure*}[h!]
    \includegraphics[width=\textwidth]{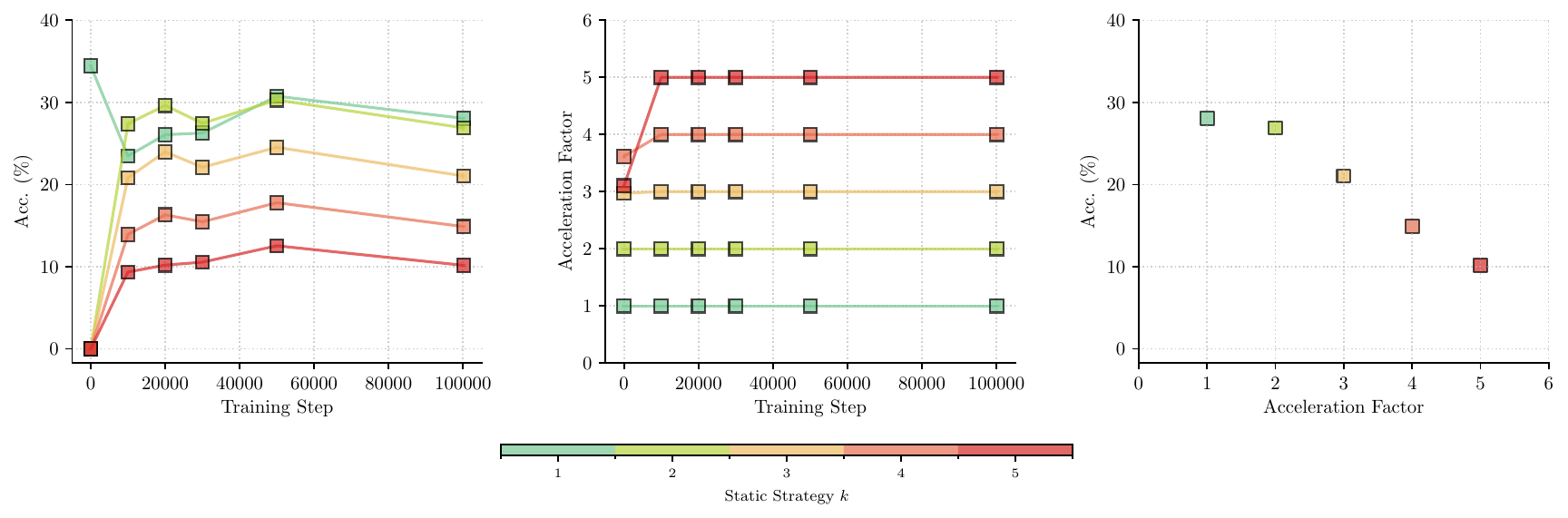}
    \includegraphics[width=\textwidth]{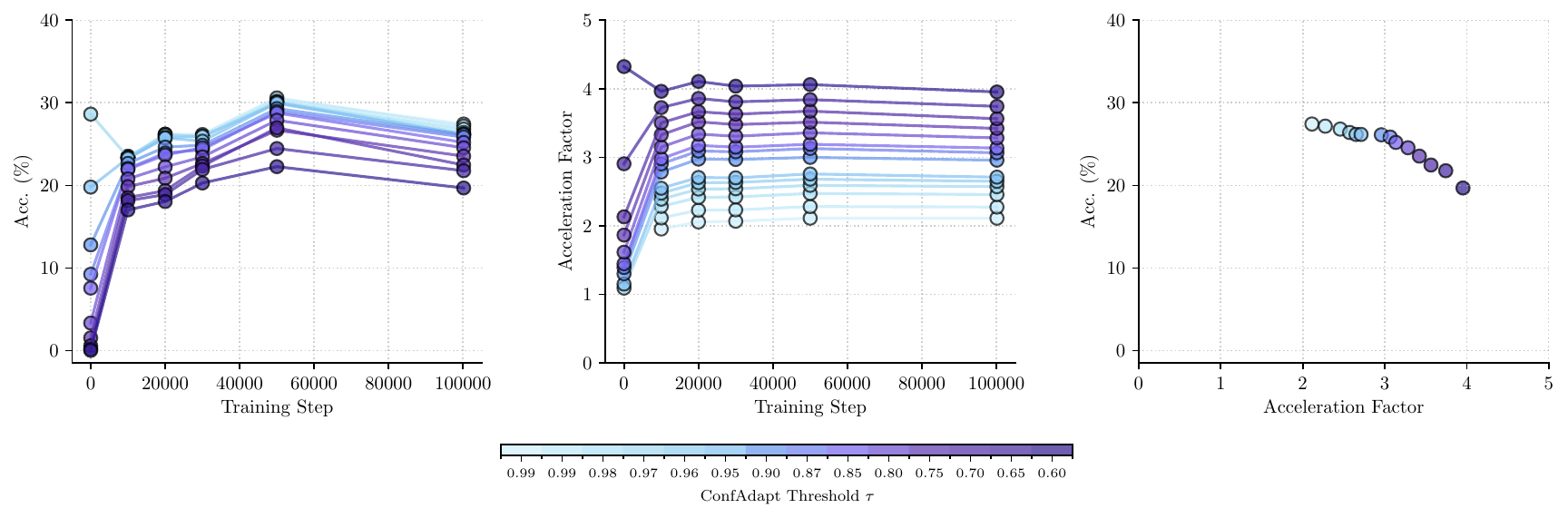}
    \caption{(Left) Accuracy and (middle) acceleration dynamics for Q3-4B-Inst-2507 evaluated on BBH over the course of training. (Right) The pareto frontier of accuracy versus acceleration based on the final model checkpoint evaluated across all decoding strategies. We find that performance and acceleration stabilize around 50k training steps, and the adaptive decoding strategies achieve pareto-optimal tradeoffs between generation speed and response quality.}
    \label{fig:dynamics-flagship-q3-bbh}
\end{figure*}

In \cref{fig:dynamics-flagship-l3-gsm,fig:dynamics-flagship-q3-gsm} we present the evaluation performance and acceleration on the GSM8K benchmark over the course of $\sim 100,000$ training iterations for our L3.1-8B-Magpie based and Q3-4B-Inst-2507 based MTP models respectively to accompany the performance vs. acceleration pareto plots in \cref{fig:dynamics-flagship-l3-q3-gsm-scatter}. We show the same dynamics and performance vs. acceleration pareto plots for both models but for the BBH evaluation task in \cref{fig:dynamics-flagship-l3-bbh,fig:dynamics-flagship-q3-bbh,fig:dynamics-flagship-l3-q3-bbh-scatter} We find that when evaluating the models under the \textit{Static} scheme at various values of $k$, performance is inversely proportional to the value of $k$. We also observe that both performance and acceleration stabilize by the halfway point. Note that we report an initial acceleration factor $< k$ for Static schemes in some cases because when untrained, the model often emits a stop token within the first $k$ token window, truncating the sequence.

The \textit{ConfAdapt} scheme is visualized on the same axes as the Static scheme by computing the average $k$ value that was chosen by the heuristic across all decoding steps over all the generations in the evaluation. We present the results across a sweep of confidence thresholds for each model and benchmark. We observe that the ConfAdapt schemes are pareto optimal, achieving significant acceleration factors of $>3\times$ without degrading accuracy very much. At more aggressive settings using more permissive confidence thresholds, the acceleration factors are as high as $5\times$ but cause a more significant reduction in accuracy.


\clearpage

\subsection{Throughput vs. Latency}\label{app:tpt-vs-latency}

We analyze the real world efficiency of our MTP approach using a prototype integration with the SGLang serving engine. The key features that this integration provides are continuous batching, compiled computation graphs, and tensor parallelism support. The specific version of SGLang we have integrated our method with is the \href{https://github.com/sgl-project/sglang}{https://github.com/sgl-project/sglang} repo at commit hash \href{https://github.com/sgl-project/sglang/commit/7b0fb43c7a7e5a6d8584e9ebdd5bc6b12e2f0fcd}{7b0fb43}.

\begin{figure}[h!]
    \includegraphics[width=0.49\textwidth, trim=0cm 0cm 0cm 0cm, clip]{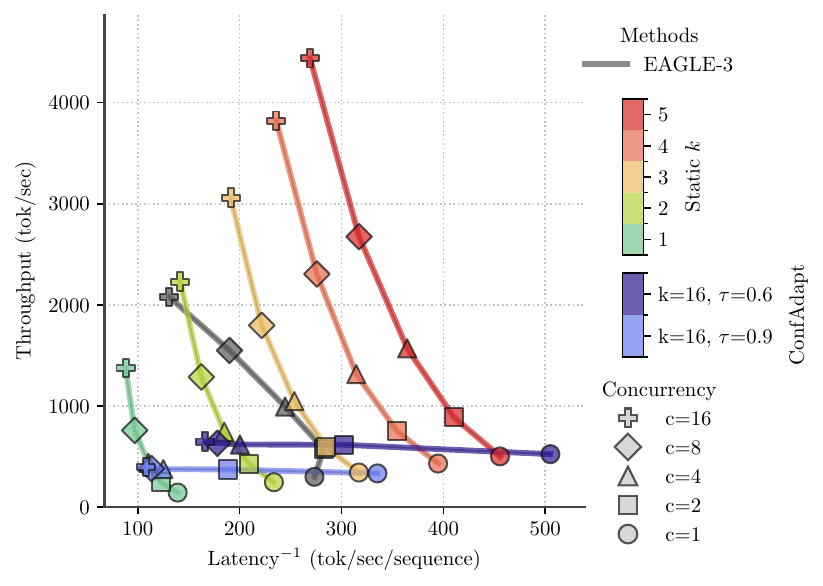}
    \includegraphics[width=0.49\textwidth, trim=0cm 0cm 0cm 0cm, clip]{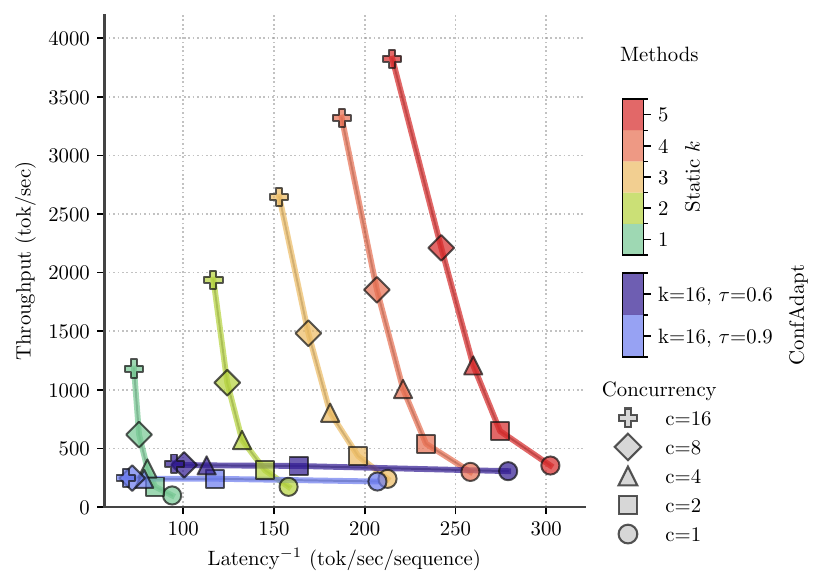}
    \caption{Throughput vs. latency tradeoff for \textbf{(Left)} L3.1-8B-Magpie model and for \textbf{(Right)} a Qwen 32B model. Throughput measures the total tokens emitted by the server per second and latency measures the tokens per second on a per request basis (i.e. per prompt or ``user''). The EAGLE-3 speculative decoding technique is shown for an 8B architecture for which a pre-trained set of speculator weights exists. Note that the 8B is run on a single GPU and the 32B is run on 4 GPUs with tensor parallelism. Concurrency is the number of requests sent to the server in parallel. \textbf{We see that static MTP decoding smoothly trades latency for throughput. However, while it is competitive with static k=3 at c=1, the per-token overhead of ConfAdapt decoding limits its concurrent scaling, at least in our prototype implementation.} Note that the irregularity for EAGLE-3 between $c=1$ and $c=2$ is likely caused by a change in engine behavior behind the scenes when switching between single prompt and batching. We observed the same thing in the curve for standard NTP decoding without EAGLE-3, though that setting is omitted to avoid cluttering an already dense chart.}
    \label{fig:tpt-pareto-8b-32b-full}
\end{figure}

\paragraph{Benchmarking as a function of concurrency.}

To test the performance of our MTP LM at varying concurrency levels, we utilize a workflow where we first start an instance of the SGLang server, and then query it using the LM Evaluation Harness' SGLang integration with varying numbers of concurrent requests. 

Specifically, for the MTP methods, the SGLang server is started with \texttt{--cuda-graph-max-bs=16} \texttt{--max-running-requests=16}, and for the EAGLE-3 server it is started with \texttt{--cuda-graph-max-bs=64} \texttt{--max-running-requests=48}. 48 appears to be an enforced hard override for this method within the version of SGLang we use (also see note below specifically for EAGLE-3). For both methods, \texttt{--dtype=bfloat16}, \texttt{--attention-backend=flashinfer}, and \texttt{--mem-fraction-static=0.70} are set. Because the prototype implementation of our MTP method does not support it at this time, and also because its support for speculative decoding methods like EAGLE-3 is also unstable in the SGLang version we use, \texttt{--disable-overlap-schedule} is set for both methods meaning overlap scheduling is disabled during all tests.

For the LM Eval Harness client, we use the \texttt{lm\_eval --model sglang-generate} backend, with the \texttt{base\_url} parameter pointing to our instantiated SGLang server. Then we use the client to send the server the first $512$ samples of the same GSM8K test set multiple times, varying the number of questions it simultaneously sends to the server by sweeping the \texttt{num\_concurrent} argument in $[1,2,4,8,16]$.

To produce \cref{fig:tpt-pareto-8b-reduced,fig:tpt-pareto-8b-32b-full}, throughput and latency are recovered based on the server logs produced by SGLang. The throughput, or total tokens per second output by the server, is measured based on a regex applied to the server's stdout log looking for the \texttt{Decode batch...gen throughput (token/s):...} pattern to extract the ``(tok/sec)'' metric. The latency, or tokens per second per request as perceived by the client, is measured by ensuring we run the server with the \texttt{--enable-metrics --export-metrics-to-file} args, and then parsing the \texttt{sglang-request-metrics} log file to retrieve the \texttt{completion\_tokens} and \texttt{e2e\_latency} fields, dividing them to get the ``(toks/sec/sequence)'' metric. For both measurements, the leading $10\%$ of the observations are trimmed to exclude any warm-up effects for the server during each client run at each concurrency level.

\paragraph{EAGLE-3 details.} We benchmark against EAGLE-3 using the official integration with SGLang. 

\href{https://docs.sglang.io/advanced\_features/speculative\_decoding.html#eagle-3-decoding}{\texttt{docs.sglang.io/advanced\_features/speculative\_decoding.html\#eagle-3-decoding}}

The specific evaluation settings are the default parameters suggested in the SGL documentation: 
\begin{itemize}[topsep=0.0cm,itemsep=-0.1cm,leftmargin=0.5cm,font=\bfseries]
\setlength{\parskip}{0.1cm}
    \item base model = \texttt{meta-llama/Meta-Llama-3.1-8B-Instruct}
    \item draft model = \texttt{jamesliu1/sglang-EAGLE3-Llama-3.1-Instruct-8B}
    \item speculative steps = 3
    \item top-k = 4
    \item draft tokens per step = 16
\end{itemize}
We note that while our evaluation setup is capped at 16 way concurrency on the client side for both MTP and EAGLE-3, for the latter, to avoid any throttling of the baseline method, as stated above, the max batch size and the running requests are allowed to be much higher for the speculative decoding method. While we believe that the public SGLang integration of EAGLE-3 is relatively well optimized, there are a range of hyperparameters and the draft model to choose, and we expect that there exist more performant combinations of parameters. It is also possible that closed source, forked versions of the implementation might exist and perform better. However, we reiterate that \cref{fig:tpt-pareto-8b-reduced,fig:tpt-pareto-8b-32b-full} are really only meant to show general ``competitiveness'' rather than any claim of superiority.

\paragraph{Implementation details for Static $k$ strategy.}
To support our static $k$ MTP decoding, we instantiate the server with the $k$ value and the max batch size stated previously to pre-compile cudagraphs. Note that the specific underlying target for this $k$ argument is actually the ``query length''. In autoregressive decoding for a NTP model using KV caching, while the sequence dimension of the matrix of key vector grows as a function of generation step, the query vector is always singleton; thus the computation is parallelized over the key dimension. For our static approach, the parallelization dim remains the same, but, for hidden states of size $d$, the query vector $(1,d)$ is now a query matrix of size $(2k-1,d)$. As discussed in \cref{sec:impl-kvcache}, we predict $k$ new tokens using k-1 masks, while we recompute and commit KV states for the previously predicted k tokens, giving us an ``active'' region, or query length compilation target, of $2k-1$.

\paragraph{Implementation details for ConfAdapt strategy.}
For the ConfAdapt scheme, server preparation is slightly more complex. We instead provide the max $k$ value and graphs are compiled for a \textit{range} of inference shapes. For some generations, the number of accepted tokens is large, and for some it is small, so we must compile graphs for query lengths in the range $(k,2k-1)$ to cover the extremal query length cases of accepting just 1 token, all the way up to accepting the the max number of tokens $k$. This variability also presents one additional issue. The SGLang cudagraphs are constructed and identified based on the query length and batch size (query length, number of sequences being processed simultaneously) with the freely variable dimension being the key matrix length to accommodate prompts/generations of different lengths. To naively accommodate the additional variability in the query length per sequence in this setting, we would also need to compile a new graph for every possible set of query lengths for each sequence, per batch size. As each graph consumes memory at runtime, even for modest concurrency levels and max values of $k$, this would cause a combinatorial increase in memory overhead. 

To ``address'' this problem, we instead limit the pre-compilation logic to only the cross of unique query lengths and batch sizes, and enforce a homogeneous execution pattern. During batched execution, the server is limited to only scheduling batches of sequences whose current query lengths are all the same; the count of which is dynamically determined by the number of tokens accepted during the previous iteration for each sequence. This means that at any time, if there is only one sequence with the query length $q$ ready, then regardless of the concurrency level---max simultaneous requests the server could theoretically try and execute---it will only be able to process that sequence in isolation. Combined with the fact that the ConfAdapt method predicts more tokens than it accepts (eg. $16$ vs $\sim3$) which saps GPU memory bandwidth and compute, the loss of the ability to make progress on $c$ sequences per setup in parallel is the most likely explanation for the poor throughput of the ConfAdapt approach at high concurrency. Thus, while we conclude that the single user, single sequence ($c=1$) performance of the ConfAdapt scheme is genuinely competitive with both static $k=3$ and EAGLE-3, the prototype implementation we present would benefit from more expert improvement and integration by SGLang's maintainers or other systems software researchers (hence the use of scare quotes around ``address'' at the start of this paragraph).

\paragraph{Testing on a larger model.}

The tensor parallelism features of SGLang also allow is to serve a larger 32B model and observe whether this introduces any differences in scaling across strategies. We train a version of Qwen 32B found on the Hugging Face hub hosted as \texttt{qywu/Qwen3-32B-Instruct} using the same MetaMathQA data and hyperparameters as our main models but with increased FSDP based model parallelism. Then we evaluate it using the same set of decoding strategies but with 4 way tensor parallelism active in SGLang. We observe almost the same scaling trends across strategies for this model with minor differences likely caused by the increased per layer, per token costs of the larger model.

\clearpage

\subsection{Performance across tasks}
We also evaluate the same two models from \cref{fig:dynamics-flagship-l3-q3-gsm-scatter} on a more general set of evaluations, and summarize the performance of both models on both benchmarks in \cref{tab:math-evals} and \cref{tab:general-evals}. Here we report the accuracy and acceleration metrics (``Effective $k$'' value) for the final checkpoint in the training run while also including the performance of each model at step 0 evaluated under standard NTP decoding (the pretrained initial checkpoint under the static $k=1$ scheme). 

We observe that even though both models were trained on only MetaMathQA, transfer learning occurs. For the L3.1-8B-Magpie model, in many cases the ConfAdapt scheme also achieves $>3\times$ acceleration with minor impact on performance relative to $k=1$ decoding. When compared to the performance of the original model checkpoints at step 0 (``Baseline''), the impact on accuracy appears more severe, but as these are highly optimized post-trained models that we finetune on a small dataset for a limited number of iterations, we think that this decrease even under $k=1$ decoding is unsurprising. 

As the two initial models have different strengths represented by performance at step 0 before MTP adaptation, we generally observe that larger average acceleration factors occur on the GSM8K and BBH benchmarks, and that for the more open-ended generative tasks like CNN DailyMail, relatively low acceleration factors are yielded for all adaptive decoding configurations.

\begin{table}[h!]
\small
\centering
\begin{tabular}{llcccccc}
\toprule
 &  & \multicolumn{2}{c}{GSM8K} & \multicolumn{2}{c}{AIME25} & \multicolumn{2}{c}{GPQA Main} \\
 &  & Acc. (\%) & Eff. k & Acc. (\%) & Eff. k & Acc. (\%) & Eff. k \\
\midrule
L3.1-8B-Magpie & Baseline Step 0, k=1 & 69.5 ± 1.3 & 1 & 0.0 ± 0.0 & 1 & 17.6 ± 1.8 & 1 \\
 Chat Template: On & Static k=1 & 66.0 ± 1.3 & 1 & 0.0 ± 0.0 & 1 & 15.4 ± 1.7 & 1 \\
 & Static k=2 & 60.7 ± 1.3 & 2 & 0.0 ± 0.0 & 2 & 16.1 ± 1.7 & 2 \\
 & Static k=3 & 53.0 ± 1.4 & 3 & 0.0 ± 0.0 & 3 & 11.4 ± 1.5 & 3 \\
 & Static k=4 & 49.8 ± 1.4 & 4 & 0.0 ± 0.0 & 4 & 8.7 ± 1.3 & 4 \\
 & Static k=5 & 41.5 ± 1.4 & 5 & 0.0 ± 0.0 & 5 & 6.5 ± 1.2 & 5 \\
 & ConfAdapt ($\tau=0.995$) & 65.7 ± 1.3 & 1.9 ± 0.5 & 0.0 ± 0.0 & 3.0 ± 1.0 & 14.7 ± 1.7 & 1.5 ± 0.3 \\
 & ConfAdapt ($\tau=0.99$) & 66.0 ± 1.3 & 2.1 ± 0.6 & 0.0 ± 0.0 & 3.6 ± 1.4 & 14.3 ± 1.7 & 1.6 ± 0.3 \\
 & ConfAdapt ($\tau=0.98$) & 65.5 ± 1.3 & 2.3 ± 0.7 & 0.0 ± 0.0 & 4.3 ± 1.5 & 14.5 ± 1.7 & 1.7 ± 0.4 \\
 & ConfAdapt ($\tau=0.97$) & 65.3 ± 1.3 & 2.5 ± 0.8 & 0.0 ± 0.0 & 4.5 ± 1.8 & 14.3 ± 1.7 & 1.7 ± 0.4 \\
 & ConfAdapt ($\tau=0.96$) & 64.9 ± 1.3 & 2.6 ± 0.8 & 0.0 ± 0.0 & 4.7 ± 1.9 & 14.3 ± 1.7 & 1.8 ± 0.4 \\
 & ConfAdapt ($\tau=0.95$) & 64.7 ± 1.3 & 2.8 ± 0.9 & 0.0 ± 0.0 & 4.8 ± 2.0 & 15.2 ± 1.7 & 1.8 ± 0.4 \\
 & ConfAdapt ($\tau=0.9$) & 64.1 ± 1.3 & 3.3 ± 1.1 & 0.0 ± 0.0 & 5.5 ± 2.4 & 15.0 ± 1.7 & 2.0 ± 0.5 \\
 & ConfAdapt ($\tau=0.87$) & 62.3 ± 1.3 & 3.6 ± 1.2 & 0.0 ± 0.0 & 5.1 ± 2.2 & 14.7 ± 1.7 & 2.1 ± 0.5 \\
 & ConfAdapt ($\tau=0.85$) & 61.2 ± 1.3 & 3.7 ± 1.3 & 0.0 ± 0.0 & 5.2 ± 2.6 & 15.6 ± 1.7 & 2.2 ± 0.5 \\
 & ConfAdapt ($\tau=0.8$) & 59.2 ± 1.4 & 4.2 ± 1.5 & 0.0 ± 0.0 & 5.8 ± 2.6 & 15.6 ± 1.7 & 2.4 ± 0.6 \\
 & ConfAdapt ($\tau=0.75$) & 56.9 ± 1.4 & 4.6 ± 1.6 & 0.0 ± 0.0 & 5.7 ± 2.7 & 14.1 ± 1.6 & 2.5 ± 0.6 \\
 & ConfAdapt ($\tau=0.7$) & 55.0 ± 1.4 & 5.1 ± 1.8 & 0.0 ± 0.0 & 6.4 ± 3.2 & 17.2 ± 1.8 & 2.7 ± 0.6 \\
 & ConfAdapt ($\tau=0.65$) & 51.5 ± 1.4 & 5.6 ± 1.9 & 0.0 ± 0.0 & 5.7 ± 3.0 & 15.0 ± 1.7 & 2.9 ± 0.7 \\
 & ConfAdapt ($\tau=0.6$) & 46.6 ± 1.4 & 6.1 ± 2.1 & 0.0 ± 0.0 & 6.4 ± 3.3 & 14.1 ± 1.6 & 3.1 ± 0.7 \\
\cline{1-8}
Q3-4B-Inst-2507 & Baseline Step 0, k=1 & 75.4 ± 1.2 & 1 & 46.7 ± 9.3 & 1 & 12.3 ± 1.6 & 1 \\
 Chat Template: Off & Static k=1 & 89.1 ± 0.9 & 1 & 23.3 ± 7.9 & 1 & 15.8 ± 1.7 & 1 \\
 & Static k=2 & 82.3 ± 1.1 & 2 & 0.0 ± 0.0 & 2 & 18.3 ± 1.8 & 2 \\
 & Static k=3 & 61.6 ± 1.3 & 3 & 0.0 ± 0.0 & 3 & 7.1 ± 1.2 & 3 \\
 & Static k=4 & 44.2 ± 1.4 & 4 & 0.0 ± 0.0 & 4 & 6.0 ± 1.1 & 4 \\
 & Static k=5 & 28.9 ± 1.2 & 5 & 0.0 ± 0.0 & 5 & 2.7 ± 0.8 & 5 \\
 & ConfAdapt ($\tau=0.995$) & 88.2 ± 0.9 & 2.1 ± 0.4 & 26.7 ± 8.2 & 2.3 ± 0.4 & 15.0 ± 1.7 & 1.2 ± 0.1 \\
 & ConfAdapt ($\tau=0.99$) & 87.6 ± 0.9 & 2.2 ± 0.5 & 13.3 ± 6.3 & 2.5 ± 0.5 & 14.1 ± 1.6 & 1.2 ± 0.1 \\
 & ConfAdapt ($\tau=0.98$) & 87.5 ± 0.9 & 2.4 ± 0.6 & 20.0 ± 7.4 & 2.6 ± 0.5 & 13.4 ± 1.6 & 1.3 ± 0.1 \\
 & ConfAdapt ($\tau=0.97$) & 87.5 ± 0.9 & 2.5 ± 0.6 & 13.3 ± 6.3 & 2.6 ± 0.5 & 13.6 ± 1.6 & 1.3 ± 0.1 \\
 & ConfAdapt ($\tau=0.96$) & 87.0 ± 0.9 & 2.6 ± 0.7 & 16.7 ± 6.9 & 2.6 ± 0.6 & 13.2 ± 1.6 & 1.3 ± 0.1 \\
 & ConfAdapt ($\tau=0.95$) & 86.9 ± 0.9 & 2.7 ± 0.7 & 23.3 ± 7.9 & 2.8 ± 0.5 & 13.6 ± 1.6 & 1.3 ± 0.2 \\
 & ConfAdapt ($\tau=0.9$) & 83.6 ± 1.0 & 3.1 ± 0.8 & 10.0 ± 5.6 & 3.1 ± 0.7 & 14.7 ± 1.7 & 1.4 ± 0.2 \\
 & ConfAdapt ($\tau=0.87$) & 82.3 ± 1.1 & 3.2 ± 0.9 & 10.0 ± 5.6 & 3.2 ± 0.5 & 15.4 ± 1.7 & 1.5 ± 0.2 \\
 & ConfAdapt ($\tau=0.85$) & 81.3 ± 1.1 & 3.3 ± 0.9 & 6.7 ± 4.6 & 3.3 ± 0.6 & 14.3 ± 1.7 & 1.5 ± 0.2 \\
 & ConfAdapt ($\tau=0.8$) & 77.6 ± 1.1 & 3.6 ± 1.0 & 10.0 ± 5.6 & 3.4 ± 0.7 & 17.2 ± 1.8 & 1.6 ± 0.2 \\
 & ConfAdapt ($\tau=0.75$) & 73.2 ± 1.2 & 3.8 ± 1.1 & 6.7 ± 4.6 & 3.6 ± 0.9 & 15.8 ± 1.7 & 1.7 ± 0.3 \\
 & ConfAdapt ($\tau=0.7$) & 68.2 ± 1.3 & 4.1 ± 1.2 & 6.7 ± 4.6 & 3.4 ± 0.9 & 14.5 ± 1.7 & 1.8 ± 0.3 \\
 & ConfAdapt ($\tau=0.65$) & 62.0 ± 1.3 & 4.4 ± 1.3 & 3.3 ± 3.3 & 3.8 ± 1.4 & 15.8 ± 1.7 & 1.8 ± 0.3 \\
 & ConfAdapt ($\tau=0.6$) & 58.8 ± 1.4 & 4.6 ± 1.3 & 6.7 ± 4.6 & 4.2 ± 1.4 & 18.1 ± 1.8 & 2.0 ± 0.3 \\
\cline{1-8}
\bottomrule
\end{tabular}
\caption{Complete evaluation results for the two main models on the math and STEM centric benchmarks.}\label{tab:math-evals}
\end{table}

\begin{table}[h!]
\small
\centering
\begin{tabular}{llcccccc}
\toprule
 &  & \multicolumn{2}{c}{BBH} & \multicolumn{2}{c}{IFEval} & \multicolumn{2}{c}{CNN DailyMail} \\
 &  & Acc. (\%) & Eff. k & Acc. (\%) & Eff. k & ROUGE-L & Eff. k \\
\midrule
L3.1-8B-Magpie & Baseline Step 0, k=1 & 67.1 ± 0.5 & 1 & 55.1 ± 2.1 & 1 & 0.226 ± 0.001 & 1 \\
 Chat Template: On & Static k=1 & 57.6 ± 0.5 & 1 & 42.5 ± 2.1 & 1 & 0.236 ± 0.001 & 1 \\
 & Static k=2 & 51.4 ± 0.5 & 2 & 29.8 ± 2.0 & 2 & 0.226 ± 0.001 & 2 \\
 & Static k=3 & 45.1 ± 0.5 & 3 & 23.5 ± 1.8 & 3 & 0.206 ± 0.001 & 3 \\
 & Static k=4 & 28.0 ± 0.5 & 4 & 22.0 ± 1.8 & 4 & 0.176 ± 0.001 & 4 \\
 & Static k=5 & 16.6 ± 0.4 & 5 & 19.8 ± 1.7 & 5 & 0.149 ± 0.001 & 5 \\
 & ConfAdapt ($\tau=0.995$) & 57.3 ± 0.5 & 2.4 ± 0.9 & 42.0 ± 2.1 & 1.5 ± 0.7 & 0.243 ± 0.001 & 1.5 ± 0.4 \\
 & ConfAdapt ($\tau=0.99$) & 56.8 ± 0.5 & 2.6 ± 1.0 & 41.8 ± 2.1 & 1.6 ± 0.8 & 0.243 ± 0.001 & 1.6 ± 0.5 \\
 & ConfAdapt ($\tau=0.98$) & 56.4 ± 0.5 & 3.0 ± 1.2 & 42.1 ± 2.1 & 1.8 ± 1.0 & 0.242 ± 0.001 & 1.7 ± 0.6 \\
 & ConfAdapt ($\tau=0.97$) & 56.2 ± 0.5 & 3.2 ± 1.3 & 42.5 ± 2.1 & 1.8 ± 1.1 & 0.242 ± 0.001 & 1.8 ± 0.6 \\
 & ConfAdapt ($\tau=0.96$) & 56.0 ± 0.5 & 3.3 ± 1.4 & 40.9 ± 2.1 & 1.9 ± 1.1 & 0.242 ± 0.001 & 1.8 ± 0.7 \\
 & ConfAdapt ($\tau=0.95$) & 55.6 ± 0.5 & 3.4 ± 1.4 & 41.4 ± 2.1 & 2.0 ± 1.2 & 0.242 ± 0.001 & 1.9 ± 0.7 \\
 & ConfAdapt ($\tau=0.9$) & 53.8 ± 0.5 & 3.9 ± 1.6 & 40.3 ± 2.1 & 2.2 ± 1.3 & 0.241 ± 0.001 & 2.1 ± 0.8 \\
 & ConfAdapt ($\tau=0.87$) & 51.9 ± 0.5 & 4.1 ± 1.7 & 40.5 ± 2.1 & 2.3 ± 1.4 & 0.241 ± 0.001 & 2.2 ± 0.8 \\
 & ConfAdapt ($\tau=0.85$) & 50.4 ± 0.5 & 4.2 ± 1.7 & 39.7 ± 2.1 & 2.3 ± 1.5 & 0.24 ± 0.001 & 2.2 ± 0.9 \\
 & ConfAdapt ($\tau=0.8$) & 45.5 ± 0.5 & 4.4 ± 1.8 & 38.6 ± 2.1 & 2.4 ± 1.5 & 0.239 ± 0.001 & 2.4 ± 0.9 \\
 & ConfAdapt ($\tau=0.75$) & 39.4 ± 0.5 & 4.6 ± 1.9 & 33.8 ± 2.0 & 2.6 ± 1.6 & 0.237 ± 0.001 & 2.5 ± 0.9 \\
 & ConfAdapt ($\tau=0.7$) & 33.6 ± 0.5 & 4.8 ± 2.0 & 35.5 ± 2.1 & 2.9 ± 1.9 & 0.236 ± 0.001 & 2.7 ± 1.0 \\
 & ConfAdapt ($\tau=0.65$) & 27.6 ± 0.5 & 5.1 ± 2.1 & 34.0 ± 2.0 & 3.1 ± 2.0 & 0.232 ± 0.001 & 2.9 ± 1.0 \\
 & ConfAdapt ($\tau=0.6$) & 21.2 ± 0.5 & 5.3 ± 2.2 & 32.2 ± 2.0 & 3.2 ± 2.0 & 0.229 ± 0.001 & 3.1 ± 1.0 \\
\cline{1-8}
Q3-4B-Inst-2507 & Baseline Step 0, k=1 & 34.4 ± 0.4 & 1 & 65.6 ± 2.0 & 1 & 0.223 ± 0.001 & 1 \\
 Chat Template: Off & Static k=1 & 28.0 ± 0.4 & 1 & 66.2 ± 2.0 & 1 & 0.227 ± 0.001 & 1 \\
 & Static k=2 & 26.9 ± 0.4 & 2 & 42.9 ± 2.1 & 2 & 0.222 ± 0.001 & 2 \\
 & Static k=3 & 21.0 ± 0.4 & 3 & 31.1 ± 2.0 & 3 & 0.205 ± 0.001 & 3 \\
 & Static k=4 & 14.9 ± 0.4 & 4 & 26.6 ± 1.9 & 4 & 0.177 ± 0.001 & 4 \\
 & Static k=5 & 10.2 ± 0.3 & 5 & 22.6 ± 1.8 & 5 & 0.156 ± 0.001 & 5 \\
 & ConfAdapt ($\tau=0.995$) & 27.4 ± 0.3 & 2.1 ± 0.8 & 64.7 ± 2.1 & 1.5 ± 0.5 & 0.235 ± 0.001 & 1.2 ± 0.1 \\
 & ConfAdapt ($\tau=0.99$) & 27.2 ± 0.3 & 2.3 ± 0.9 & 64.3 ± 2.1 & 1.5 ± 0.5 & 0.235 ± 0.001 & 1.2 ± 0.1 \\
 & ConfAdapt ($\tau=0.98$) & 26.8 ± 0.4 & 2.5 ± 1.0 & 62.3 ± 2.1 & 1.6 ± 0.6 & 0.234 ± 0.001 & 1.2 ± 0.1 \\
 & ConfAdapt ($\tau=0.97$) & 26.4 ± 0.4 & 2.6 ± 1.1 & 61.9 ± 2.1 & 1.7 ± 0.6 & 0.234 ± 0.001 & 1.3 ± 0.1 \\
 & ConfAdapt ($\tau=0.96$) & 26.2 ± 0.4 & 2.6 ± 1.1 & 62.1 ± 2.1 & 1.7 ± 0.7 & 0.234 ± 0.001 & 1.3 ± 0.1 \\
 & ConfAdapt ($\tau=0.95$) & 26.2 ± 0.4 & 2.7 ± 1.1 & 61.6 ± 2.1 & 1.7 ± 0.7 & 0.234 ± 0.001 & 1.3 ± 0.1 \\
 & ConfAdapt ($\tau=0.9$) & 26.1 ± 0.4 & 3.0 ± 1.2 & 59.5 ± 2.1 & 1.9 ± 0.9 & 0.233 ± 0.001 & 1.4 ± 0.1 \\
 & ConfAdapt ($\tau=0.87$) & 25.8 ± 0.4 & 3.1 ± 1.3 & 58.2 ± 2.1 & 2.0 ± 0.9 & 0.232 ± 0.001 & 1.4 ± 0.2 \\
 & ConfAdapt ($\tau=0.85$) & 25.2 ± 0.4 & 3.1 ± 1.3 & 59.3 ± 2.1 & 2.1 ± 1.0 & 0.232 ± 0.001 & 1.5 ± 0.2 \\
 & ConfAdapt ($\tau=0.8$) & 24.6 ± 0.4 & 3.3 ± 1.4 & 55.5 ± 2.1 & 2.2 ± 1.0 & 0.231 ± 0.001 & 1.5 ± 0.2 \\
 & ConfAdapt ($\tau=0.75$) & 23.5 ± 0.4 & 3.4 ± 1.4 & 55.3 ± 2.1 & 2.3 ± 1.1 & 0.23 ± 0.001 & 1.6 ± 0.2 \\
 & ConfAdapt ($\tau=0.7$) & 22.5 ± 0.4 & 3.6 ± 1.4 & 56.7 ± 2.1 & 2.5 ± 1.2 & 0.229 ± 0.001 & 1.7 ± 0.2 \\
 & ConfAdapt ($\tau=0.65$) & 21.8 ± 0.4 & 3.7 ± 1.5 & 51.8 ± 2.2 & 2.7 ± 1.3 & 0.228 ± 0.001 & 1.8 ± 0.3 \\
 & ConfAdapt ($\tau=0.6$) & 19.7 ± 0.4 & 4.0 ± 1.5 & 49.9 ± 2.2 & 3.0 ± 1.6 & 0.227 ± 0.001 & 1.9 ± 0.3 \\
\cline{1-8}
\bottomrule
\end{tabular}
\caption{Complete evaluation results for the two main models on the instruction following and open ended generation benchmarks.}\label{tab:general-evals}
\end{table}

\clearpage

\subsection{Ablations}\label{app:ablations}

\subsection{Alternate finetuning dataset.}\label{app:abl-data}
We also evaluate a version of the L3.1-8B-Magpie model further finetuned on the Magpie dataset instead of the MetaMathQA data used in all other experiments to analyze the impact of minimizing the distribution shift between the online MTP objective and the most recent stage of training for the initial checkpoint we use. In \cref{tab:abl-l3-data-evals} we can see that while, the $k=1$ performance of the models is quite similar, and there are a few instances of decoding strategy and task pairings where the model trained on Magpie performs better, in general, the version trained on MetaMathQA achieves better accuracy and acceleration in most cases. We suspect that this is due to the fact that the Magpie datasets are much larger and therefore at the same iteration count

\begin{table}[h!]
\small
\centering
\begin{tabular}{llcccccc}
\toprule
 &  & \multicolumn{2}{c}{GSM8K} & \multicolumn{2}{c}{BBH} & \multicolumn{2}{c}{IFEval} \\
 &  & Acc. (\%) & Eff. k & Acc. (\%) & Eff. k & Acc. (\%) & Eff. k \\
\midrule
L3.1-8B-Magpie & Baseline Step 0, k=1 & 69.5 ± 1.3 & 1 & 67.1 ± 0.5 & 1 & 55.1 ± 2.1 & 1 \\
 Chat Template: On & Static k=1 & 66.0 ± 1.3 & 1 & 57.6 ± 0.5 & 1 & 42.5 ± 2.1 & 1 \\
 & Static k=2 & 60.7 ± 1.3 & 2 & 51.4 ± 0.5 & 2 & 29.8 ± 2.0 & 2 \\
 & Static k=3 & 53.0 ± 1.4 & 3 & 45.1 ± 0.5 & 3 & 23.5 ± 1.8 & 3 \\
 & Static k=4 & 49.8 ± 1.4 & 4 & 28.0 ± 0.5 & 4 & 22.0 ± 1.8 & 4 \\
 & Static k=5 & 41.5 ± 1.4 & 5 & 16.6 ± 0.4 & 5 & 19.8 ± 1.7 & 5 \\
 & ConfAdapt ($\tau=0.995$) & 65.7 ± 1.3 & 1.9 ± 0.5 & 57.3 ± 0.5 & 2.4 ± 0.9 & 42.0 ± 2.1 & 1.5 ± 0.7 \\
 & ConfAdapt ($\tau=0.99$) & 66.0 ± 1.3 & 2.1 ± 0.6 & 56.8 ± 0.5 & 2.6 ± 1.0 & 41.8 ± 2.1 & 1.6 ± 0.8 \\
 & ConfAdapt ($\tau=0.98$) & 65.5 ± 1.3 & 2.3 ± 0.7 & 56.4 ± 0.5 & 3.0 ± 1.2 & 42.1 ± 2.1 & 1.8 ± 1.0 \\
 & ConfAdapt ($\tau=0.97$) & 65.3 ± 1.3 & 2.5 ± 0.8 & 56.2 ± 0.5 & 3.2 ± 1.3 & 42.5 ± 2.1 & 1.8 ± 1.1 \\
 & ConfAdapt ($\tau=0.96$) & 64.9 ± 1.3 & 2.6 ± 0.8 & 56.0 ± 0.5 & 3.3 ± 1.4 & 40.9 ± 2.1 & 1.9 ± 1.1 \\
 & ConfAdapt ($\tau=0.95$) & 64.7 ± 1.3 & 2.8 ± 0.9 & 55.6 ± 0.5 & 3.4 ± 1.4 & 41.4 ± 2.1 & 2.0 ± 1.2 \\
 & ConfAdapt ($\tau=0.9$) & 64.1 ± 1.3 & 3.3 ± 1.1 & 53.8 ± 0.5 & 3.9 ± 1.6 & 40.3 ± 2.1 & 2.2 ± 1.3 \\
 & ConfAdapt ($\tau=0.87$) & 62.3 ± 1.3 & 3.6 ± 1.2 & 51.9 ± 0.5 & 4.1 ± 1.7 & 40.5 ± 2.1 & 2.3 ± 1.4 \\
 & ConfAdapt ($\tau=0.85$) & 61.2 ± 1.3 & 3.7 ± 1.3 & 50.4 ± 0.5 & 4.2 ± 1.7 & 39.7 ± 2.1 & 2.3 ± 1.5 \\
 & ConfAdapt ($\tau=0.8$) & 59.2 ± 1.4 & 4.2 ± 1.5 & 45.5 ± 0.5 & 4.4 ± 1.8 & 38.6 ± 2.1 & 2.4 ± 1.5 \\
 & ConfAdapt ($\tau=0.75$) & 56.9 ± 1.4 & 4.6 ± 1.6 & 39.4 ± 0.5 & 4.6 ± 1.9 & 33.8 ± 2.0 & 2.6 ± 1.6 \\
 & ConfAdapt ($\tau=0.7$) & 55.0 ± 1.4 & 5.1 ± 1.8 & 33.6 ± 0.5 & 4.8 ± 2.0 & 35.5 ± 2.1 & 2.9 ± 1.9 \\
 & ConfAdapt ($\tau=0.65$) & 51.5 ± 1.4 & 5.6 ± 1.9 & 27.6 ± 0.5 & 5.1 ± 2.1 & 34.0 ± 2.0 & 3.1 ± 2.0 \\
 & ConfAdapt ($\tau=0.6$) & 46.6 ± 1.4 & 6.1 ± 2.1 & 21.2 ± 0.5 & 5.3 ± 2.2 & 32.2 ± 2.0 & 3.2 ± 2.0 \\
\cline{1-8}
L3.1-8B-Magpie FT Magpie & Baseline Step 0, k=1 & 69.5 ± 1.3 & 1 & 67.1 ± 0.5 & 1 & 55.1 ± 2.1 & 1 \\
 Chat Template: On & Static k=1 & 64.5 ± 1.3 & 1 & 57.7 ± 0.5 & 1 & 41.8 ± 2.1 & 1 \\
 & Static k=2 & 53.8 ± 1.4 & 2 & 54.2 ± 0.5 & 2 & 30.3 ± 2.0 & 2 \\
 & Static k=3 & 45.3 ± 1.4 & 3 & 49.9 ± 0.5 & 3 & 27.9 ± 1.9 & 3 \\
 & Static k=4 & 34.0 ± 1.3 & 4 & 34.2 ± 0.5 & 4 & 24.6 ± 1.9 & 4 \\
 & Static k=5 & 26.2 ± 1.2 & 5 & 24.9 ± 0.5 & 5 & 20.0 ± 1.7 & 5 \\
 & ConfAdapt ($\tau=0.995$) & 64.5 ± 1.3 & 1.5 ± 0.3 & 57.3 ± 0.5 & 2.8 ± 1.1 & 41.6 ± 2.1 & 1.5 ± 0.9 \\
 & ConfAdapt ($\tau=0.99$) & 64.6 ± 1.3 & 1.7 ± 0.3 & 54.9 ± 0.5 & 3.2 ± 1.4 & 41.8 ± 2.1 & 1.6 ± 1.1 \\
 & ConfAdapt ($\tau=0.98$) & 64.4 ± 1.3 & 1.8 ± 0.4 & 46.7 ± 0.5 & 3.7 ± 1.7 & 41.0 ± 2.1 & 1.7 ± 1.3 \\
 & ConfAdapt ($\tau=0.97$) & 64.0 ± 1.3 & 2.0 ± 0.5 & 38.6 ± 0.5 & 4.0 ± 1.9 & 41.0 ± 2.1 & 1.7 ± 1.4 \\
 & ConfAdapt ($\tau=0.96$) & 64.1 ± 1.3 & 2.1 ± 0.5 & 34.6 ± 0.5 & 4.3 ± 2.0 & 40.7 ± 2.1 & 1.8 ± 1.5 \\
 & ConfAdapt ($\tau=0.95$) & 63.5 ± 1.3 & 2.1 ± 0.6 & 31.0 ± 0.5 & 4.5 ± 2.1 & 40.5 ± 2.1 & 1.8 ± 1.5 \\
 & ConfAdapt ($\tau=0.9$) & 61.5 ± 1.3 & 2.5 ± 0.7 & 18.8 ± 0.4 & 5.1 ± 2.4 & 39.7 ± 2.1 & 2.0 ± 1.6 \\
 & ConfAdapt ($\tau=0.87$) & 60.7 ± 1.3 & 2.7 ± 0.8 & 16.8 ± 0.4 & 5.4 ± 2.5 & 39.6 ± 2.1 & 2.1 ± 1.8 \\
 & ConfAdapt ($\tau=0.85$) & 59.4 ± 1.4 & 2.8 ± 0.8 & 15.8 ± 0.4 & 5.6 ± 2.5 & 39.4 ± 2.1 & 2.1 ± 1.8 \\
 & ConfAdapt ($\tau=0.8$) & 54.9 ± 1.4 & 3.1 ± 0.9 & 14.7 ± 0.4 & 6.0 ± 2.7 & 38.8 ± 2.1 & 2.3 ± 1.9 \\
 & ConfAdapt ($\tau=0.75$) & 50.9 ± 1.4 & 3.4 ± 1.0 & 13.2 ± 0.4 & 6.3 ± 2.7 & 37.7 ± 2.1 & 2.4 ± 2.0 \\
 & ConfAdapt ($\tau=0.7$) & 47.9 ± 1.4 & 3.7 ± 1.1 & 12.3 ± 0.4 & 6.7 ± 2.8 & 36.4 ± 2.1 & 2.6 ± 2.1 \\
 & ConfAdapt ($\tau=0.65$) & 42.6 ± 1.4 & 4.0 ± 1.2 & 10.2 ± 0.3 & 7.1 ± 2.9 & 35.7 ± 2.1 & 2.9 ± 2.3 \\
 & ConfAdapt ($\tau=0.6$) & 37.4 ± 1.3 & 4.3 ± 1.3 & 8.8 ± 0.3 & 7.4 ± 3.0 & 33.5 ± 2.0 & 3.2 ± 2.5 \\
\cline{1-8}
\bottomrule
\end{tabular}
\caption{Complete evaluation results for the main L3.1-8B-Magpie model trained on MetaMathQA versus the version trained on Magpie on a subset of the evaluation benchmarks.}\label{tab:abl-l3-data-evals}
\end{table}

\clearpage

\subsection{Offline Ground Truth Supervision (``GT Suprv.'').}\label{app:abl-gt}

As a baseline, we perform an ablation where all parameters of the MTP training objective are the same as in the main experiments, but the source of supervision (target tokens) no longer comes from the teacher model's feedback. Specifically, instead of using the argmax feedback from the teacher as the labels in \cref{eq:mtpo2}, we simply treat the $k$ ground truth tokens that appeared at each MTP region $r$ in the prepared and masked input sequence as the labels. As shown in~\cref{fig:dynamics-flagship-l3-online-offline,tab:abl-l3-suprv-evals-pt1}, this underperforms the online objective at the higher values of $k$.

\subsection{Bidirectional attention in MTP region (``BDA'').}\label{app:abl-bda}

Due to the fact that the student MTP LM does not receive ground truth tokens as input within the masked region where it is performing prediction during training, our training objective does not have the potential for the ``anti-causal'' leakage that would occur in traditional offline NTP training if a \textit{bidirectional} attention mask was used. Thus, we are able to consider allowing all neighboring mask token representations within the same MTP region to attend to each other. As shown in \cref{tab:abl-l3-suprv-evals-pt1}, we find that increasing the representational and computational capacity of the MTP prediction pass does not yield a meaningfully better final model under the same training conditions.

\subsection{Prefix Loss NTP Supervision}\label{app:abl-prefix-loss}

Since we prepare input data during training using the $M=N/(2k)$ strided convention, as opposed to say $M=N/k$, certain regions in the sequence are ``prefix'' only (see ``Pred'' regions marked false/0 in \cref{fig:explainer-tok-masking}). Due to the carefully blocked attention mask, the logits at those positions in the MTP model's output are produced by a computation that exactly matches the standard NTP forward pass. As a result, we consider also supervising the standard NTP objective on all of the prefix positions but as shown in \cref{tab:abl-l3-suprv-evals-pt2}, we find that it underperforms the objective configuration used for the main experiments.

\subsection{Unique Mask Tokens}\label{app:abl-unique-masks}

One observed failure mode observed during our experiments is the repetition of tokens at the end of a MTP region for larger values of $k$. To investigate whether this was potentially the result of the model having insufficiently fine grained positional information to distinguish tokens $j$ from $j-1$ within the $k$ token region, we also train a model with $k$ unique \texttt{<MASK>} tokens rather than just one. Specifically, during training, all ask regions were implemented as \texttt{<MTP1>,<MTP2>,<MTP3>,...} and at test time, the generation loop extended the sequence before each step with the same unique mask token sequence. As shown in \cref{tab:abl-l3-suprv-evals-pt3}, this approach does not yield meaningfully better results than the main setting, even though slightly more unique trainable parameters (rows of the embedding matrix) were available to the model. 

While not shown, we also qualitatively observe that the model trained with a single mask token and the model trained with unique mask tokens, were actually rather agnostic to which masking strategy was employed at test time. In other words, the main model trained with a single mask token could accept the unique token format and operate relatively normally, and vice versa. We conclude this is evidence of the token embeddings themselves playing a rather minor role and the computational ``slot'' provided in the model to be the primary functional role they play; it seems that the positional embeddings (RoPE features) used at each mask position provides the main source of positional information to the model during MTP.

\subsection{Offline Teacher}\label{app:abl-offline-teacher}

To interrogate our claim about the superiority of the online objective posed in \cref{eq:mtpo2}, we explore an alternate type of supervision sitting somewhere between truly offline ground truth forcing and the online self-distillation approach. In this ablation, the prompts (questions) in the same set of MetaMathQA training examples used in the main results are passed to the original L3.1-8B-Magpie checkpoint to allow it to generate a CoT and answer. Then, after this offline step is complete, theses new generations are used as the targets under a similar configuration of the training code as the ``GT Suprv.'' setting; no teacher pass is performed and the students predictions are simply directly compared to the $k$ tokens that appeared in this ``offline teacher'''s roll out. 

\begin{figure*}[h!]
    \begin{subfigure}[b]{0.33\textwidth}
    \includegraphics[width=\textwidth, trim=0cm 0cm 0.0cm 0cm, clip]{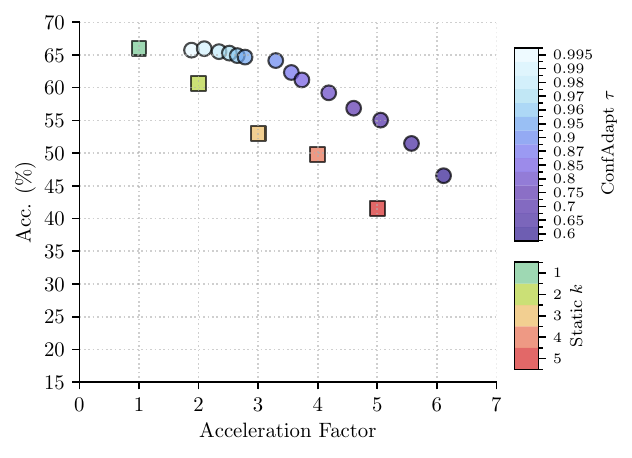}
    \caption{Online, main setting.}
    \end{subfigure}
    \begin{subfigure}[b]{0.33\textwidth}
    \includegraphics[width=\textwidth, trim=0cm 0cm 0.0cm 0cm, clip]{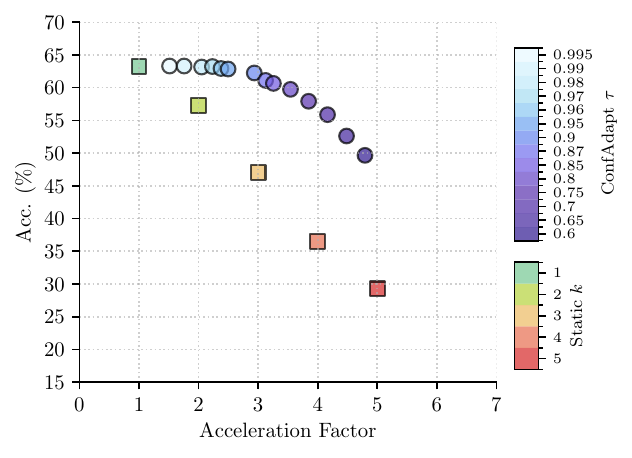}
    \caption{``Offline Teacher''}
    \end{subfigure}
    \begin{subfigure}[b]{0.33\textwidth}
    \includegraphics[width=\textwidth, trim=0cm 0cm 0.0cm 0cm, clip]{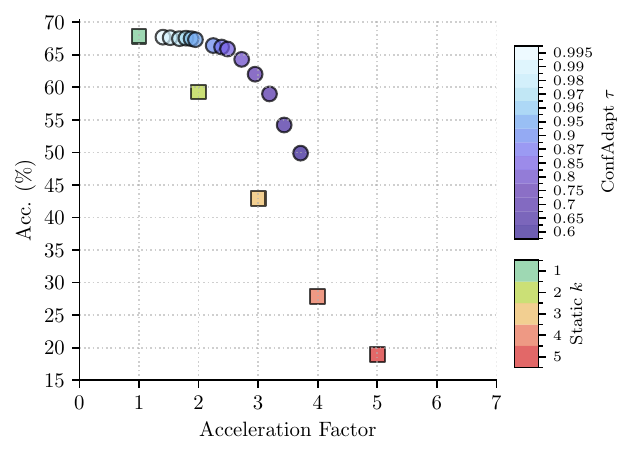}
    \caption{``GT Suprv.''}
    \end{subfigure}
    \caption{The performance of our (\textbf{Left}) L3.1-8B-Magpie based MTP LM using the main settings, equivalent to left of~\cref{fig:dynamics-flagship-l3-q3-gsm-scatter} and \cref{alg:singleshot-online-train-step}, (\textbf{Middle}) the ``Offline Teacher'' variant of the objective, and (\textbf{Right}) the ``GT Suprv.'' variant evaluated on the GSM8K benchmark after $\sim$100k steps of training. These settings are described in \cref{app:abl-offline-teacher} and performance is tabulated in \cref{tab:abl-l3-suprv-evals-pt1,tab:abl-l3-suprv-evals-pt3}. Performance tradeoff is visualized by plotting the effective $k$ value or ``Acceleration Factor'' versus the Accuracy on the benchmark; axes are shared to make lateral comparison possible. \textbf{While the performance of the two offline configurations are comparable to the main online setting up to effective $k$ of about $\sim2$, at larger values of $k$ the online method is far superior. For example, there is a $20\%$ acc. gap at static $k=5$ between the online and GT Suprv.}}
    \label{fig:dynamics-flagship-l3-online-offline}
\end{figure*}

As shown in \cref{fig:dynamics-flagship-l3-online-offline,tab:abl-l3-suprv-evals-pt3}, we observe that this approach is performs slightly better than the ``GT Suprv.'' setting, but worse than the online setting. While it is competitive with the true online setting at small values of $k$, the performance at larger values of $k$ is consistently worse. We conclude that this middle ranking between true online and true offline is intuitive given the pseudo-on-policy nature of NTP rollouts from the same set of weights we are about to tune for MTP.

\begin{figure*}[h!]
    \begin{subfigure}[b]{0.49\textwidth}
    \includegraphics[width=\textwidth, trim=0cm 0cm 0.0cm 0cm, clip]{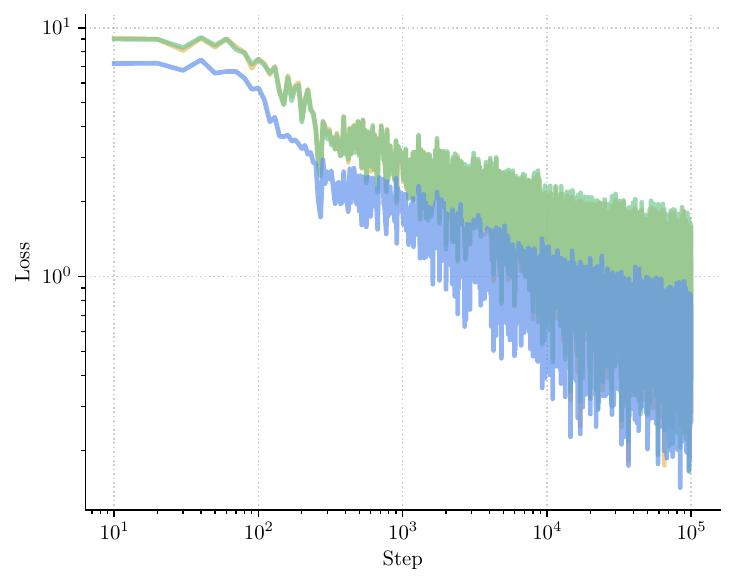}
    \caption{}
    \end{subfigure}
    \begin{subfigure}[b]{0.49\textwidth}
    \includegraphics[width=\textwidth, trim=0cm 0cm 0.0cm 0cm, clip]{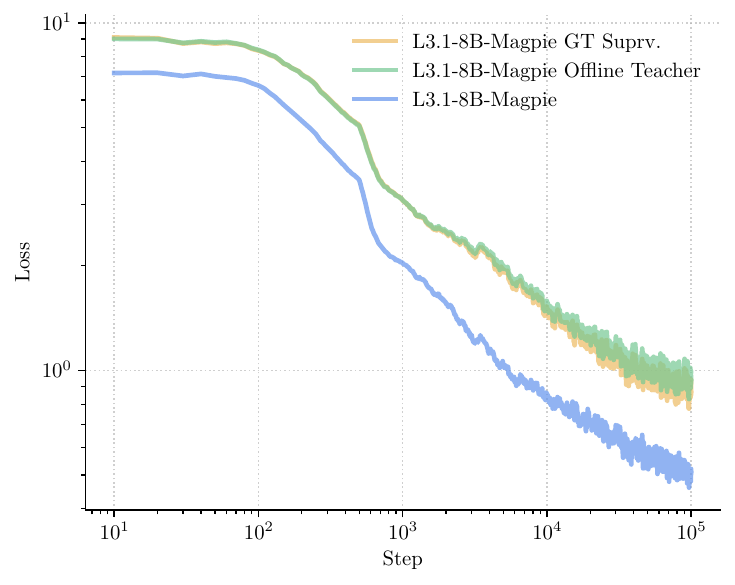}
    \caption{}
    \end{subfigure}
    \caption{Training loss as a function of optimizer step, across $\sim 100,000$ steps of training for our L3.1-8B-Magpie based MTP LM trained using the main online objective of~\cref{fig:dynamics-flagship-l3-q3-gsm-scatter} and \cref{alg:singleshot-online-train-step}, compared to the ``Offline Teacher'' and offline``GT Suprv.'' variants of the objective. We present both \textbf{(a)} the raw loss on semilog axes, and \textbf{(b)} a smoothed version via a 50 step windowed rolling average. The high per step variance is due to the $k$ value randomization in all three runs. \textbf{We see that while the lowest per step losses achieved by all three methods are similar, the highest loss values for the offline methods are much larger than those in the online setting.} This translates both to a large gap in the rolling average and corroborates the large gap in evaluation performance at higher values of $k$ shown in \cref{fig:dynamics-flagship-l3-online-offline}.}
    \label{fig:train-dynamics-flagship-l3-online-offline}
\end{figure*}

\clearpage
\begin{table}[h!]
\small
\vspace{-1.0cm}
\centering
\begin{tabular}{llcc}
\toprule
 &  & \multicolumn{2}{c}{GSM8K} \\
 &  & Acc. (\%) & Eff. k \\
\midrule
L3.1-8B-Magpie & Baseline Step 0, k=1 & 69.5 ± 1.3 & 1 \\
 Chat Template: On & Static k=1 & 66.0 ± 1.3 & 1 \\
 & Static k=2 & 60.7 ± 1.3 & 2 \\
 & Static k=3 & 53.0 ± 1.4 & 3 \\
 & Static k=4 & 49.8 ± 1.4 & 4 \\
 & Static k=5 & 41.5 ± 1.4 & 5 \\
 & ConfAdapt ($\tau=0.995$) & 65.7 ± 1.3 & 1.9 ± 0.5 \\
 & ConfAdapt ($\tau=0.99$) & 66.0 ± 1.3 & 2.1 ± 0.6 \\
 & ConfAdapt ($\tau=0.98$) & 65.5 ± 1.3 & 2.3 ± 0.7 \\
 & ConfAdapt ($\tau=0.97$) & 65.3 ± 1.3 & 2.5 ± 0.8 \\
 & ConfAdapt ($\tau=0.96$) & 64.9 ± 1.3 & 2.6 ± 0.8 \\
 & ConfAdapt ($\tau=0.95$) & 64.7 ± 1.3 & 2.8 ± 0.9 \\
 & ConfAdapt ($\tau=0.9$) & 64.1 ± 1.3 & 3.3 ± 1.1 \\
 & ConfAdapt ($\tau=0.87$) & 62.3 ± 1.3 & 3.6 ± 1.2 \\
 & ConfAdapt ($\tau=0.85$) & 61.2 ± 1.3 & 3.7 ± 1.3 \\
 & ConfAdapt ($\tau=0.8$) & 59.2 ± 1.4 & 4.2 ± 1.5 \\
 & ConfAdapt ($\tau=0.75$) & 56.9 ± 1.4 & 4.6 ± 1.6 \\
 & ConfAdapt ($\tau=0.7$) & 55.0 ± 1.4 & 5.1 ± 1.8 \\
 & ConfAdapt ($\tau=0.65$) & 51.5 ± 1.4 & 5.6 ± 1.9 \\
 & ConfAdapt ($\tau=0.6$) & 46.6 ± 1.4 & 6.1 ± 2.1 \\
\cline{1-4}
L3.1-8B-Magpie GT Suprv. & Baseline Step 0, k=1 & 69.5 ± 1.3 & 1 \\
 Chat Template: On & Static k=1 & 67.9 ± 1.3 & 1 \\
 & Static k=2 & 59.3 ± 1.4 & 2 \\
 & Static k=3 & 42.9 ± 1.4 & 3 \\
 & Static k=4 & 27.8 ± 1.2 & 4 \\
 & Static k=5 & 19.0 ± 1.1 & 5 \\
 & ConfAdapt ($\tau=0.995$) & 67.7 ± 1.3 & 1.4 ± 0.2 \\
 & ConfAdapt ($\tau=0.99$) & 67.6 ± 1.3 & 1.5 ± 0.3 \\
 & ConfAdapt ($\tau=0.98$) & 67.5 ± 1.3 & 1.7 ± 0.3 \\
 & ConfAdapt ($\tau=0.97$) & 67.6 ± 1.3 & 1.8 ± 0.4 \\
 & ConfAdapt ($\tau=0.96$) & 67.5 ± 1.3 & 1.9 ± 0.4 \\
 & ConfAdapt ($\tau=0.95$) & 67.3 ± 1.3 & 1.9 ± 0.4 \\
 & ConfAdapt ($\tau=0.9$) & 66.4 ± 1.3 & 2.2 ± 0.6 \\
 & ConfAdapt ($\tau=0.87$) & 66.2 ± 1.3 & 2.4 ± 0.6 \\
 & ConfAdapt ($\tau=0.85$) & 65.9 ± 1.3 & 2.5 ± 0.7 \\
 & ConfAdapt ($\tau=0.8$) & 64.3 ± 1.3 & 2.7 ± 0.8 \\
 & ConfAdapt ($\tau=0.75$) & 62.0 ± 1.3 & 2.9 ± 0.8 \\
 & ConfAdapt ($\tau=0.7$) & 59.0 ± 1.4 & 3.2 ± 0.9 \\
 & ConfAdapt ($\tau=0.65$) & 54.2 ± 1.4 & 3.4 ± 1.0 \\
 & ConfAdapt ($\tau=0.6$) & 49.9 ± 1.4 & 3.7 ± 1.1 \\
\cline{1-4}
L3.1-8B-Magpie BDA & Baseline Step 0, k=1 & 69.5 ± 1.3 & 1 \\
 Chat Template: On & Static k=1 & 63.5 ± 1.3 & 1 \\
 & Static k=2 & 56.0 ± 1.4 & 2 \\
 & Static k=3 & 47.9 ± 1.4 & 3 \\
 & Static k=4 & 43.1 ± 1.4 & 4 \\
 & Static k=5 & 37.2 ± 1.3 & 5 \\
 & ConfAdapt ($\tau=0.995$) & 63.4 ± 1.3 & 1.9 ± 0.5 \\
 & ConfAdapt ($\tau=0.99$) & 63.5 ± 1.3 & 2.1 ± 0.6 \\
 & ConfAdapt ($\tau=0.98$) & 63.3 ± 1.3 & 2.3 ± 0.7 \\
 & ConfAdapt ($\tau=0.97$) & 63.2 ± 1.3 & 2.4 ± 0.7 \\
 & ConfAdapt ($\tau=0.96$) & 62.8 ± 1.3 & 2.6 ± 0.8 \\
 & ConfAdapt ($\tau=0.95$) & 62.7 ± 1.3 & 2.7 ± 0.8 \\
 & ConfAdapt ($\tau=0.9$) & 59.8 ± 1.4 & 3.1 ± 1.0 \\
 & ConfAdapt ($\tau=0.87$) & 58.9 ± 1.4 & 3.3 ± 1.1 \\
 & ConfAdapt ($\tau=0.85$) & 58.5 ± 1.4 & 3.5 ± 1.2 \\
 & ConfAdapt ($\tau=0.8$) & 55.1 ± 1.4 & 3.8 ± 1.3 \\
 & ConfAdapt ($\tau=0.75$) & 52.2 ± 1.4 & 4.2 ± 1.4 \\
 & ConfAdapt ($\tau=0.7$) & 49.0 ± 1.4 & 4.6 ± 1.5 \\
 & ConfAdapt ($\tau=0.65$) & 46.0 ± 1.4 & 5.0 ± 1.6 \\
 & ConfAdapt ($\tau=0.6$) & 41.3 ± 1.4 & 5.3 ± 1.6 \\
\cline{1-4}
\bottomrule
\end{tabular}
\caption{Evaluation results for the ablations of the training objective configuration on the L3.1-8B-Magpie model measured on GSM8K. The first group labeled just \textbf{``L3.1-8B-Magpie''} is the configuration used for the main experiments, the \textbf{``GT Suprv.''} uses the ground truth tokens as labels rather than the online teacher forcing, \textbf{``BDA''} allows bidirectional attention in the MTP regions. table continues below.}\label{tab:abl-l3-suprv-evals-pt1}
\end{table}

\clearpage
\begin{table}[h!]
\small
\vspace{-1.0cm}
\centering
\begin{tabular}{llcc}
\toprule
 &  & \multicolumn{2}{c}{GSM8K} \\
 &  & Acc. (\%) & Eff. k \\
\midrule
L3.1-8B-Magpie Soft Teacher & Baseline Step 0, k=1 & 69.5 ± 1.3 & 1 \\
 Chat Template: On & Static k=1 & 63.6 ± 1.3 & 1 \\
 & Static k=2 & 56.1 ± 1.4 & 2 \\
 & Static k=3 & 49.7 ± 1.4 & 3 \\
 & Static k=4 & 45.6 ± 1.4 & 4 \\
 & Static k=5 & 39.2 ± 1.3 & 5 \\
 & ConfAdapt ($\tau=0.995$) & 63.5 ± 1.3 & 1.3 ± 0.2 \\
 & ConfAdapt ($\tau=0.99$) & 63.5 ± 1.3 & 1.5 ± 0.3 \\
 & ConfAdapt ($\tau=0.98$) & 63.6 ± 1.3 & 1.7 ± 0.4 \\
 & ConfAdapt ($\tau=0.97$) & 63.6 ± 1.3 & 1.8 ± 0.5 \\
 & ConfAdapt ($\tau=0.96$) & 63.6 ± 1.3 & 1.9 ± 0.5 \\
 & ConfAdapt ($\tau=0.95$) & 63.5 ± 1.3 & 2.0 ± 0.5 \\
 & ConfAdapt ($\tau=0.9$) & 62.5 ± 1.3 & 2.3 ± 0.7 \\
 & ConfAdapt ($\tau=0.87$) & 62.1 ± 1.3 & 2.5 ± 0.8 \\
 & ConfAdapt ($\tau=0.85$) & 62.1 ± 1.3 & 2.6 ± 0.9 \\
 & ConfAdapt ($\tau=0.8$) & 61.0 ± 1.3 & 2.9 ± 1.0 \\
 & ConfAdapt ($\tau=0.75$) & 58.9 ± 1.4 & 3.2 ± 1.2 \\
 & ConfAdapt ($\tau=0.7$) & 56.8 ± 1.4 & 3.6 ± 1.3 \\
 & ConfAdapt ($\tau=0.65$) & 53.7 ± 1.4 & 4.0 ± 1.5 \\
 & ConfAdapt ($\tau=0.6$) & 49.3 ± 1.4 & 4.5 ± 1.8 \\
\cline{1-4}
L3.1-8B-Magpie Static k=16 & Baseline Step 0, k=1 & 69.5 ± 1.3 & 1 \\
 Chat Template: On & Static k=1 & 54.1 ± 1.4 & 1 \\
 & Static k=2 & 45.6 ± 1.4 & 2 \\
 & Static k=3 & 38.7 ± 1.3 & 3 \\
 & Static k=4 & 35.1 ± 1.3 & 4 \\
 & Static k=5 & 32.4 ± 1.3 & 5 \\
 & ConfAdapt ($\tau=0.995$) & 54.1 ± 1.4 & 1.5 ± 0.3 \\
 & ConfAdapt ($\tau=0.99$) & 54.1 ± 1.4 & 1.7 ± 0.4 \\
 & ConfAdapt ($\tau=0.98$) & 54.1 ± 1.4 & 1.9 ± 0.6 \\
 & ConfAdapt ($\tau=0.97$) & 53.4 ± 1.4 & 2.0 ± 0.7 \\
 & ConfAdapt ($\tau=0.96$) & 53.2 ± 1.4 & 2.2 ± 0.8 \\
 & ConfAdapt ($\tau=0.95$) & 53.4 ± 1.4 & 2.3 ± 0.8 \\
 & ConfAdapt ($\tau=0.9$) & 51.6 ± 1.4 & 2.7 ± 1.1 \\
 & ConfAdapt ($\tau=0.87$) & 50.6 ± 1.4 & 3.0 ± 1.2 \\
 & ConfAdapt ($\tau=0.85$) & 50.6 ± 1.4 & 3.1 ± 1.3 \\
 & ConfAdapt ($\tau=0.8$) & 48.7 ± 1.4 & 3.5 ± 1.5 \\
 & ConfAdapt ($\tau=0.75$) & 45.9 ± 1.4 & 3.9 ± 1.6 \\
 & ConfAdapt ($\tau=0.7$) & 44.0 ± 1.4 & 4.3 ± 1.8 \\
 & ConfAdapt ($\tau=0.65$) & 41.5 ± 1.4 & 4.8 ± 2.0 \\
 & ConfAdapt ($\tau=0.6$) & 39.0 ± 1.3 & 5.3 ± 2.2 \\
\cline{1-4}
L3.1-8B-Magpie Prefix Loss & Baseline Step 0, k=1 & 69.5 ± 1.3 & 1 \\
 Chat Template: On & Static k=1 & 66.6 ± 1.3 & 1 \\
 & Static k=2 & 57.8 ± 1.4 & 2 \\
 & Static k=3 & 47.3 ± 1.4 & 3 \\
 & Static k=4 & 45.8 ± 1.4 & 4 \\
 & Static k=5 & 37.5 ± 1.3 & 5 \\
 & ConfAdapt ($\tau=0.995$) & 69.4 ± 1.3 & 1.6 ± 0.3 \\
 & ConfAdapt ($\tau=0.99$) & 69.5 ± 1.3 & 1.8 ± 0.4 \\
 & ConfAdapt ($\tau=0.98$) & 69.6 ± 1.3 & 2.0 ± 0.6 \\
 & ConfAdapt ($\tau=0.97$) & 69.3 ± 1.3 & 2.2 ± 0.7 \\
 & ConfAdapt ($\tau=0.96$) & 69.4 ± 1.3 & 2.3 ± 0.8 \\
 & ConfAdapt ($\tau=0.95$) & 69.3 ± 1.3 & 2.4 ± 0.8 \\
 & ConfAdapt ($\tau=0.9$) & 64.6 ± 1.3 & 2.7 ± 0.9 \\
 & ConfAdapt ($\tau=0.87$) & 66.3 ± 1.3 & 3.1 ± 1.1 \\
 & ConfAdapt ($\tau=0.85$) & 65.7 ± 1.3 & 3.3 ± 1.2 \\
 & ConfAdapt ($\tau=0.8$) & 63.2 ± 1.3 & 3.6 ± 1.3 \\
 & ConfAdapt ($\tau=0.75$) & 58.3 ± 1.4 & 4.0 ± 1.5 \\
 & ConfAdapt ($\tau=0.7$) & 54.4 ± 1.4 & 4.4 ± 1.6 \\
 & ConfAdapt ($\tau=0.65$) & 50.9 ± 1.4 & 4.9 ± 1.8 \\
 & ConfAdapt ($\tau=0.6$) & 47.5 ± 1.4 & 5.0 ± 1.7 \\
\cline{1-4}
\bottomrule
\end{tabular}
\caption{Table continues from above. The \textbf{``Soft Teacher''} is trained using the full teacher logits rather than just the argmax as labels, \textbf{``Static $k$''} keeps $k$ fixed at the max value of 16 throughout training, and \textbf{``Prefix Loss''} includes the auxiliary loss term supervising the standard NTP loss on the non-MTP positions in each sequence.}\label{tab:abl-l3-suprv-evals-pt2}
\end{table}

\clearpage
\begin{table}[h!]
\small
\vspace{-1.0cm}
\centering
\begin{tabular}{llcc}
\toprule
 &  & \multicolumn{2}{c}{GSM8K} \\
 &  & Acc. (\%) & Eff. k \\
\midrule
L3.1-8B-Magpie Static k=9 & Baseline Step 0, k=1 & 69.5 ± 1.3 & 1 \\
 Chat Template: On & Static k=1 & 60.8 ± 1.3 & 1 \\
 & Static k=2 & 54.1 ± 1.4 & 2 \\
 & Static k=3 & 51.6 ± 1.4 & 3 \\
 & Static k=4 & 43.7 ± 1.4 & 4 \\
 & Static k=5 & 41.1 ± 1.4 & 5 \\
 & ConfAdapt ($\tau=0.995$) & 60.6 ± 1.3 & 1.8 ± 0.5 \\
 & ConfAdapt ($\tau=0.99$) & 60.5 ± 1.3 & 2.0 ± 0.6 \\
 & ConfAdapt ($\tau=0.98$) & 60.6 ± 1.3 & 2.3 ± 0.7 \\
 & ConfAdapt ($\tau=0.97$) & 60.5 ± 1.3 & 2.4 ± 0.8 \\
 & ConfAdapt ($\tau=0.96$) & 60.4 ± 1.3 & 2.6 ± 0.9 \\
 & ConfAdapt ($\tau=0.95$) & 60.2 ± 1.3 & 2.7 ± 0.9 \\
 & ConfAdapt ($\tau=0.9$) & 58.5 ± 1.4 & 3.2 ± 1.1 \\
 & ConfAdapt ($\tau=0.87$) & 57.9 ± 1.4 & 3.5 ± 1.2 \\
 & ConfAdapt ($\tau=0.85$) & 56.7 ± 1.4 & 3.6 ± 1.3 \\
 & ConfAdapt ($\tau=0.8$) & 54.2 ± 1.4 & 4.0 ± 1.4 \\
 & ConfAdapt ($\tau=0.75$) & 52.8 ± 1.4 & 4.4 ± 1.6 \\
 & ConfAdapt ($\tau=0.7$) & 49.6 ± 1.4 & 4.8 ± 1.7 \\
 & ConfAdapt ($\tau=0.65$) & 47.4 ± 1.4 & 5.3 ± 1.7 \\
 & ConfAdapt ($\tau=0.6$) & 41.5 ± 1.4 & 5.7 ± 1.9 \\
\cline{1-4}
L3.1-8B-Magpie Unique Mask Tokens & Baseline Step 0, k=1 & 69.5 ± 1.3 & 1 \\
 Chat Template: On & Static k=1 & 66.7 ± 1.3 & 1 \\
 & Static k=2 & 58.8 ± 1.4 & 2 \\
 & Static k=3 & 52.2 ± 1.4 & 3 \\
 & Static k=4 & 46.7 ± 1.4 & 4 \\
 & Static k=5 & 40.3 ± 1.4 & 5 \\
 & ConfAdapt ($\tau=0.995$) & 66.7 ± 1.3 & 1.8 ± 0.5 \\
 & ConfAdapt ($\tau=0.99$) & 66.7 ± 1.3 & 2.0 ± 0.6 \\
 & ConfAdapt ($\tau=0.98$) & 66.5 ± 1.3 & 2.3 ± 0.7 \\
 & ConfAdapt ($\tau=0.97$) & 66.2 ± 1.3 & 2.4 ± 0.8 \\
 & ConfAdapt ($\tau=0.96$) & 65.7 ± 1.3 & 2.6 ± 0.9 \\
 & ConfAdapt ($\tau=0.95$) & 65.6 ± 1.3 & 2.7 ± 0.9 \\
 & ConfAdapt ($\tau=0.9$) & 64.0 ± 1.3 & 3.2 ± 1.1 \\
 & ConfAdapt ($\tau=0.87$) & 63.1 ± 1.3 & 3.5 ± 1.3 \\
 & ConfAdapt ($\tau=0.85$) & 62.1 ± 1.3 & 3.6 ± 1.3 \\
 & ConfAdapt ($\tau=0.8$) & 59.7 ± 1.4 & 4.1 ± 1.5 \\
 & ConfAdapt ($\tau=0.75$) & 57.5 ± 1.4 & 4.5 ± 1.6 \\
 & ConfAdapt ($\tau=0.7$) & 54.6 ± 1.4 & 4.9 ± 1.8 \\
 & ConfAdapt ($\tau=0.65$) & 49.7 ± 1.4 & 5.4 ± 2.0 \\
 & ConfAdapt ($\tau=0.6$) & 46.0 ± 1.4 & 5.9 ± 2.1 \\
\cline{1-4}
L3.1-8B-Magpie Offline Teacher & Baseline Step 0, k=1 & 69.5 ± 1.3 & 1 \\
 Chat Template: On & Static k=1 & 63.2 ± 1.3 & 1 \\
 & Static k=2 & 57.2 ± 1.4 & 2 \\
 & Static k=3 & 47.0 ± 1.4 & 3 \\
 & Static k=4 & 36.5 ± 1.3 & 4 \\
 & Static k=5 & 29.3 ± 1.3 & 5 \\
 & ConfAdapt ($\tau=0.995$) & 63.3 ± 1.3 & 1.5 ± 0.2 \\
 & ConfAdapt ($\tau=0.99$) & 63.3 ± 1.3 & 1.8 ± 0.4 \\
 & ConfAdapt ($\tau=0.98$) & 63.2 ± 1.3 & 2.1 ± 0.5 \\
 & ConfAdapt ($\tau=0.97$) & 63.2 ± 1.3 & 2.2 ± 0.6 \\
 & ConfAdapt ($\tau=0.96$) & 62.9 ± 1.3 & 2.4 ± 0.7 \\
 & ConfAdapt ($\tau=0.95$) & 62.9 ± 1.3 & 2.5 ± 0.7 \\
 & ConfAdapt ($\tau=0.9$) & 62.2 ± 1.3 & 2.9 ± 0.9 \\
 & ConfAdapt ($\tau=0.87$) & 61.1 ± 1.3 & 3.1 ± 1.0 \\
 & ConfAdapt ($\tau=0.85$) & 60.7 ± 1.3 & 3.3 ± 1.0 \\
 & ConfAdapt ($\tau=0.8$) & 59.7 ± 1.4 & 3.5 ± 1.1 \\
 & ConfAdapt ($\tau=0.75$) & 57.9 ± 1.4 & 3.8 ± 1.2 \\
 & ConfAdapt ($\tau=0.7$) & 55.9 ± 1.4 & 4.2 ± 1.3 \\
 & ConfAdapt ($\tau=0.65$) & 52.6 ± 1.4 & 4.5 ± 1.4 \\
 & ConfAdapt ($\tau=0.6$) & 49.7 ± 1.4 & 4.8 ± 1.6 \\
\cline{1-4}
\bottomrule
\end{tabular}
\caption{Table continues from above. The \textbf{``Static $\mathbf{k=9}$''} run is an additional static ablation that keeps $k$ fixed at approximately the average value the $k \in (2,16)$ would see during training, \textbf{``Unique Mask Tokens''} is trained and evaluated with \texttt{<MTP1>,<MTP2>,<MTP3>,...} in the masked regions rather than just the repeated \texttt{<MTP1>}, and \textbf{``Offline Teacher''} is similar to the ``GT Suprv.'' run but the targets were generated using the teacher checkpoint itself prior to training.}\label{tab:abl-l3-suprv-evals-pt3}
\end{table}

\end{document}